\documentclass[11pt]{article}

\usepackage[preprint]{acl}

\usepackage{times}
\usepackage{latexsym}

\usepackage[T1]{fontenc}

\usepackage[utf8]{inputenc}

\usepackage{microtype}

\IfFileExists{inconsolata.sty}{\usepackage{inconsolata}}{}

\usepackage{amsmath, amssymb, amsfonts, mathtools}

\usepackage{graphicx}
\usepackage{hyperref}

\usepackage{CJK}

\usepackage{pifont} 
\usepackage{booktabs}
\usepackage{multirow}
\usepackage{makecell}
\usepackage{adjustbox}
\usepackage{enumitem}
\usepackage{xspace}
\usepackage{colortbl}
\usepackage{array}

\usepackage{natbib}
\usepackage{url}

\usepackage{subcaption}
\usepackage{chngcntr} 
\usepackage{longtable}
\usepackage{tikz}
\usetikzlibrary{shapes.geometric, arrows.meta, positioning, fit, calc,
                backgrounds, decorations.pathreplacing}

\usepackage[most]{tcolorbox}

\usepackage{float}
\floatstyle{ruled}
\newfloat{algorithm}{tbph}{loa}
\floatname{algorithm}{Algorithm}

\definecolor{tabhead}{RGB}{231,237,243}
\definecolor{tabalt}{RGB}{248,250,252}
\definecolor{accentA}{RGB}{34,77,131}
\definecolor{accentSys}{RGB}{93,52,148}
\definecolor{accentEvi}{RGB}{30,109,114}
\definecolor{accentTQwen}{RGB}{200,90,40}
\definecolor{accentTDS}{RGB}{160,80,160}
\definecolor{accentTLlama}{RGB}{50,120,80}
\definecolor{accentCons}{RGB}{34,107,68}
\definecolor{accentChosen}{RGB}{34,107,68}
\definecolor{accentRej}{RGB}{170,30,30}
\definecolor{accentStuOK}{RGB}{34,77,131}
\definecolor{accentStuBad}{RGB}{170,30,30}
\definecolor{accentSample}{RGB}{90,90,110}
\definecolor{accentRubric}{RGB}{120,80,30}

\newtcolorbox{tracebox}[1][]{%
  enhanced, breakable, width=\linewidth, colback=white,
  colframe=black!15, boxrule=0.25pt,
  borderline west={1.4pt}{0pt}{accentA},
  arc=1pt, boxsep=2pt, left=4pt,right=3pt,top=3pt,bottom=3pt,
  fontupper=\footnotesize,
  before upper={\sloppy\emergencystretch=2em},
  fonttitle=\footnotesize\itshape, coltitle=accentA,
  colbacktitle=white, titlerule=0pt, #1
}

\newtcolorbox{calloutbox}[2][]{%
  enhanced, breakable, width=\linewidth, colback=white,
  colframe=black!15, boxrule=0.25pt,
  borderline west={1.4pt}{0pt}{#2},
  arc=1pt, boxsep=2pt, left=4pt,right=3pt,top=3pt,bottom=3pt,
  fontupper=\footnotesize,
  before upper={\sloppy\emergencystretch=2em},
  fonttitle=\footnotesize\itshape, coltitle=#2,
  colbacktitle=white, titlerule=0pt, #1
}

\newcommand{\iotitlefont}{\fontfamily{phv}\fontseries{b}\fontshape{n}\selectfont}

\newtcolorbox{iocontract}[1][]{%
  enhanced, unbreakable, width=\linewidth,
  colback=white, colframe=accentSys!40,
  colbacktitle=accentSys, coltitle=white,
  boxrule=0.4pt, titlerule=0pt, arc=2.5pt, outer arc=2.5pt,
  borderline west={1.6pt}{0pt}{accentSys},
  boxsep=1pt, left=7pt, right=7pt, top=3pt, bottom=3pt,
  toptitle=2pt, bottomtitle=2pt, lefttitle=7pt, righttitle=7pt,
  fontupper=\footnotesize,
  parskip=1pt,
  before upper={\sloppy\emergencystretch=2em
                \setlength{\parskip}{1pt}},
  fonttitle=\footnotesize\iotitlefont,
  #1
}

\newcommand{\iohead}[2]{%
  \par\addvspace{1pt}\noindent
  {\setlength{\fboxsep}{2pt}%
   \colorbox{#1}{\color{white}\iotitlefont\scriptsize\,\textls[100]{#2}\,}}%
  \hspace{4pt}\ignorespaces
}

\newcommand{\iohi}[2]{{\color{#1!75!black}\textbf{#2}}}

\newtcblisting{codebox}[1][]{%
  enhanced, breakable, width=\linewidth, colback=black!1,
  colframe=black!15, boxrule=0.25pt,
  borderline west={1.4pt}{0pt}{accentSample},
  arc=1pt, boxsep=1.5pt, left=3pt,right=2pt,top=2pt,bottom=2pt,
  fonttitle=\footnotesize\itshape, coltitle=accentSample,
  colbacktitle=white, titlerule=0pt, listing only,
  listing options={
    basicstyle=\ttfamily\scriptsize,
    breaklines=true, breakatwhitespace=false,
    columns=fullflexible, keepspaces=true,
    showstringspaces=false, upquote=true,
    aboveskip=0pt, belowskip=0pt,
  }, #1
}

\DeclareUrlCommand{\code}{\urlstyle{tt}}

\newcommand{\fA}{\ensuremath{f_{\textsc{fam}}}}
\newcommand{\fS}{\ensuremath{f_{\textsc{sub}}}}
\newcommand{\fD}{\ensuremath{f_{\textsc{dir}}}}
\newcommand{\Pair}{\ensuremath{\mathrm{Pair}}}
\newcommand{\AB}{\ensuremath{\textsc{ab}}}
\newcommand{\BA}{\ensuremath{\textsc{ba}}}
\newcommand{\PRM}{\ensuremath{{\Phi_{\textsc{prm}}}}}
\newcommand{\Loss}{\ensuremath{\mathcal{L}}}
\newcommand{\Ours}{\textsc{MARD}\xspace}
\newcommand{\OursTab}{\textbf{MARD-7B}}

\newcommand{\OursSeven}{\textsc{MARD-7B}\xspace}
\newcommand{\OursITSTab}{\textbf{MARD-7B + ITS}}

\title{MARD: Mirror-Augmented Reasoning Distillation for Mechanism-Level Drug–Drug Interaction Prediction}

\author{
Mohammadreza Riyazat\textsuperscript{1},
Vian Lelo\textsuperscript{1},
Rameen Jafri\textsuperscript{1},
Yumna Khan\textsuperscript{1},
Abeer Badawi\textsuperscript{2,3} \\
\textsuperscript{1}University of Guelph, Canada \\
\textsuperscript{2}York University, Canada \\
\textsuperscript{3}Vector Institute, Canada \\
\texttt{\{mriyazat,vlelo,rjafri,ykhan04\}@uoguelph.ca},
\texttt{abeerbadawi@yorku.ca}
}

\hypersetup{
  colorlinks=true,
  citecolor=accentA,
  linkcolor=black,
  urlcolor=accentA,
  filecolor=accentA,
}

\begin{document}
\maketitle

\begin{abstract}
Mechanism-level drug-drug interaction (DDI) prediction requires identifying which enzyme or pharmacodynamic axis is implicated, in which direction, and with which evidence -- not merely whether two drugs interact. We introduce a reproducible mechanism-level DDI labelling and evaluation protocol with a structured 7-family/147-subtype taxonomy, leakage-safe cold-split protocols, and auditable reasoning metrics for evaluating pharmacological prediction beyond flat interaction classification. We propose a pipeline that produces a 7B reasoning \Ours (Mirror-Augmented Reasoning Distillation), combining three training innovations: a single-token KL divergence on direction tag that ties the model's prediction, per-loss PRM-weighted DPO with programmatic hard negatives, and a leakage-safe mechanism-aware retrieval channel. Process-reward step labels are automatically verifiable against DrugBank-structured fields, requiring no human or LLM judges. On the April-2026 DrugBank release, our \OursSeven is the only system in a $32$-system comparison whose accuracy survives drug-pair novelty, beating the best baseline by $+13.9$\,pp and GPT-4o by $+6.7$\,pp at $\sim\!1\%$ of frontier API cost. Further analysis reveals an anti-memorisation signature where accuracy improves on rarely seen drugs, suggesting that gain comes from structured pharmacological reasoning rather than drug-frequency memorisation. We release corpus, DDI-PRM, retrieval index, and training code \footnote{
Anonymous repository:
\url{https://anonymous.4open.science/r/ddi-prm-verifier-E49E/}
}.
\end{abstract}


\section{Introduction}
\label{sec:intro}

Adverse drug-drug interactions cause an estimated $5$--$10\%$ of hospital admissions~\citep{Komagamine2024}, a burden compounded by the fact that polypharmacy affects 52\% of inpatients globally~\citep{Kim2024}. The clinical question is not ``does A interact with B'' but the mechanistic one: \emph{which} enzyme, transporter, or pharmacodynamic axis is implicated, in \emph{which} direction the interaction proceeds, and \emph{which evidence} supports the claim. Unsupported ``possible interaction'' alerts drive alert fatigue, and are routinely dismissed. A systematic review and meta-analysis of 16 studies found that prescribers override 90\% of DDI alerts generated by clinical decision support systems~\citep{Felisberto2024}. The root cause is not alert volume but content: a flag of \emph{possible interaction} without a specified mechanism, direction, or evidence base gives the clinician nothing actionable to act on. Yet almost every existing benchmark reduces DDI to top-1 label accuracy of a feature- or graph-based classifier~\citep{ryu2018deepddi,deng2020multimodal,yu2021sumgnn}, and recent rationale-emitting systems~\citep{sun2025exddi,zhu2024zeroddi} neither constrain the rationale to an auditable schema nor enforce the pharmacological symmetry that flipping a pair must mirror its prediction.

Frontier reasoning LLMs that could in principle supply such rationales fail along three orthogonal axes when applied directly: \emph{(i)~mirror inconsistency} -- the same pair presented as $(A,B)$ vs.\ $(B,A)$ yields different family/direction predictions in $51.4\%$ of cases for an imitation baseline; \emph{(ii)~class-imbalance collapse} -- seven mechanism families span a $47{\times}$ size ratio and a naive cross-entropy \Ours (Mirror-Augmented Reasoning Distillation) collapses onto \textsc{AdverseRisk} attractor; \emph{(iii)~evidence hallucination} -- trained to imitate teacher full-traces parrots phrasing without grounding, citing phantom CYP flags or non-existent protein targets that a pharmacist cannot verify.

\paragraph{Research question.} Can a Small Language Model (SLM), with no test-time access to a frontier model, produce DDI predictions that are jointly \emph{schema-grounded}, \emph{mirror-stable}, \emph{robust to cold drugs and pairs} (drugs and combinations unseen at training time), and \emph{auditable} (every citation traceable to a finite, structured evidence pool)?

\paragraph{Approach.} We frame the task as constrained reasoning distillation. The resulting system, \textbf{\Ours} (\textbf{M}irror-%
\textbf{A}ugmented \textbf{R}easoning \textbf{D}istillation), is a $7$B student that receives a query pair and a structured evidence pool and emits a verifiable schema-constrained structured trace. Four stages address (i)--(iii) jointly -- cross-teacher consensus
(\S\ref{sec:consensus}), mirror-augmented SFT with position-restricted symmetry-KL (\S\ref{sec:mirror_aug}), PRM-weighted DPO with hard
negatives (\S\ref{sec:dpo}), and mechanism-aware retrieval (\S\ref{sec:retrieval}) -- over the April-2026 DrugBank release
($1.45$M pairs, $7$ families $\times$ $147$ subtypes) under three split protocols we created: \textsc{random-split (warm)}, \textsc{drug-cold}, and \textsc{pair-cold}.

\paragraph{Headline Results.} \OursSeven{} is the only system in a $32$-system comparison whose accuracy survives drug-pair novelty
(+13.9\,pp over the strongest baseline (DDIMDL), $+6.7$\,pp over GPT-4o on \textsc{pair-cold} at $\sim$$1\%$ of API cost) while remaining mirror-stable
(\textbf{Mirror Family Stability} (MFS)$\!\ge\!0.97$ and \textbf{Mirror Prediction Symmetry} (MPS)$\!\ge\!0.78$), near hallucination-free
(\textbf{Hallucination Rate} (HR)$\!=\!3.7\!\times\!10^{-4}$), and citation-grounded through strong
\textbf{Context-Support Alignment} (CSA). In summary, our work presents four contributions:
\begin{itemize}[leftmargin=*,nosep,topsep=2pt]
\item \textbf{A reproducible mechanism-level DDI labelling and protocol.} A $7$-family $\times\,147$-subtype taxonomy with directionality, three leakage-safe splits (\textsc{warm}, \textsc{drug-cold}, \textsc{pair-cold}), and four trace-quality metrics (MFS, MPS, CSA, HR), released as code and DrugBank-ID manifests for licensed reproduction.
\item \textbf{\Ours{}: a coupled training recipe.} Three new ingredients --- position-restricted symmetry-KL at the direction-tag
token (an inductive bias for structured LLM prediction), a PRM trained on auto-verifiable DrugBank step labels, and programmatic hard negatives confined to the structured \texttt{final\_answer} block.

\item \textbf{Judge-free training signal for clinical reasoning.} Schema-grounded DDI traces are deterministically verifiable against DrugBank, turning process rewards and preference data into auto-labelled signals with no human raters or LLM judges in the loop.

\item \textbf{Evidence that the win is reasoning, not memorisation.}
Across structural, open-medical, and frontier baselines, \Ours{} is the only system whose per-decile accuracy \emph{rises} on rarely-seen drugs -- the sign flip that separates pharmacological reasoning from drug-fingerprint look-up.
\end{itemize}

\section{Related Work}
\label{sec:related}

\paragraph{Predictive DDI.}
The dominant line treats DDI as supervised multi-label classification over a flat label space: \textsc{DeepDDI}~\citep{ryu2018deepddi}, \textsc{DDIMDL}~\citep{deng2020multimodal}, \textsc{CASTER}~\citep{huang2020caster}, \textsc{SumGNN}~\citep{yu2021sumgnn}, \textsc{DSN-DDI}~\citep{li2023dsnddi}, \textsc{LaGAT}~\citep{hong2022lagat}, and \textsc{MRCGNN}~\citep{xiong2023mrcgnn}, packaged into the $20$-system \textsc{OpenDDI} suite~\citep{openddi2024}. These systems predict a single flat family label and provide no subtype, direction, rationale, or symmetry guarantee.

\paragraph{Rationale-emitting and retrieval-augmented DDI.}
A smaller line targets the rationale rather than the label. \textsc{ExDDI}~\citep{sun2025exddi} retrieves DrugBank descriptions and asks an instruction-tuned LLM for a free-form explanation evaluated only at the label level. \textsc{ZeroDDI}~\citep{zhu2024zeroddi} composes biological semantics for zero-shot inductive prediction. The closest prior work is \textsc{CBR-DDI}~\citep{cbrddi2025}, which retrieves historical-case pairs through hybrid semantic-and-structural similarity and reports an ablation showing the retrieved block is load-bearing. However, as a prompting-only framework over a frozen LLM, it does not isolate the retrieval contribution under a \emph{fixed fine-tuned} model on the \emph{same} pairs. Direct prompting of frontier and medical LLMs has been shown to perform below specialised baselines on DDI~\citep{devito2025phi,singhal2023largescale}, motivating distillation over zero-shot use.

\paragraph{Reasoning distillation, process rewards, and symmetry.}
Step-wise reasoning distillation~\citep{magister2023teaching,fu2023specialising,chen2025skipthinking} with outcome- or process-reward verifiers~\citep{cobbe2021gsm8k,uesato2022solving,lightman2023letsverify} is mature in mathematical reasoning, and PRM-weighted DPO with per-step gradient weighting has been demonstrated by \textsc{Full-Step-DPO}~\citep{xu2025fullstepdpo} and \textsc{R-PRM}~\citep{she2025rprm}. We share the per-loss PRM-weighting mechanism but instantiate it for clinical mechanism prediction, where step labels are auto-verifiable against DrugBank structured fields~\citep{knox2024drugbank}. Symmetry under input reordering is a known weakness of autoregressive LLMs~\citep{berglund2024reversal,chen2024premise}; prior responses apply sequence-level consistency objectives over full output~\citep{kumar2022consistency,hejabi2025flipflop}. Our position-restricted symmetry-KL targets a single direction-tag token, an inductive bias not, to our knowledge, used before for structured LLM prediction. Self-consistency~\citep{wang2022sc}, best-of-$N$ reranking~\citep{cobbe2021gsm8k}, and conformal selective prediction~\citep{vovk2005algorithmic,romano2020classification} are standard test-time methods that we compose on a trained student. 


\section{Methodology}
\label{sec:method}
We pose drug--drug interaction (DDI) mechanism prediction as constrained reasoning over a structured evidence pool: given a pair of drugs, predict the interaction \emph{family}, \emph{subtype}, and \emph{direction} with a rationale citing only verifiable evidence IDs. After fixing the task, schema, and mirror requirement (\S\ref{sec:task_and_data}--\S\ref{sec:notation}), the method develops five components: cross-teacher consensus distillation (\S\ref{sec:consensus}), a fine-tuned PRM with auto-verifiable step labels (\S\ref{sec:prm}), mirror-augmented SFT with position-restricted symmetry-KL (\S\ref{sec:mirror_aug}), PRM-weighted DPO (\S\ref{sec:dpo}), and a leakage-safe mechanism-aware retrieval channel (\S\ref{sec:retrieval}). A glossary of all acronyms and short-form terms is provided in Appendix~\ref{app:acronyms}.

\begin{iocontract}[title={\textls[40]{Worked example: one drug pair, end-to-end}}]
\iohead{accentEvi}{INPUT}
A = \emph{Voriconazole} (antifungal),\\ B = \emph{Axitinib} (kinase-inhibitor anticancer). \\For each drug, the model receives a fixed schema of pharmacological fields (metabolic enzymes, transporters, molecular targets, pathways), four pair-level similarity scores, and a short list of labelled neighbour pairs.
\iohead{accentCons}{OUTPUT}
A JSON document with a $3$--$8$-step reasoning trace -- each step cites specific input facts and states which drug acts on which -- and a structured prediction (family, subtype, direction, polarity, confidence, abstain flag, one-sentence summary). \iohi{accentCons}{For this pair: Voriconazole inhibits mediated metabolism of Axitinib, raising Axitinib exposure.}
\iohead{accentRubric}{CONTRACT}
The gold answer never appears in the prompt; because the input is structured, every reasoning step is checkable from the data alone. \iohi{accentRubric}{We can verify citations, step--prediction consistency, and hallucinations without a downstream LLM judge.} This is what lets us train the PRM with auto-verifiable labels (\S\ref{sec:prm}) and report citation-support and hallucination rates.
\end{iocontract}

\begin{figure*}[t]
  \centering
  \includegraphics[width=\textwidth]{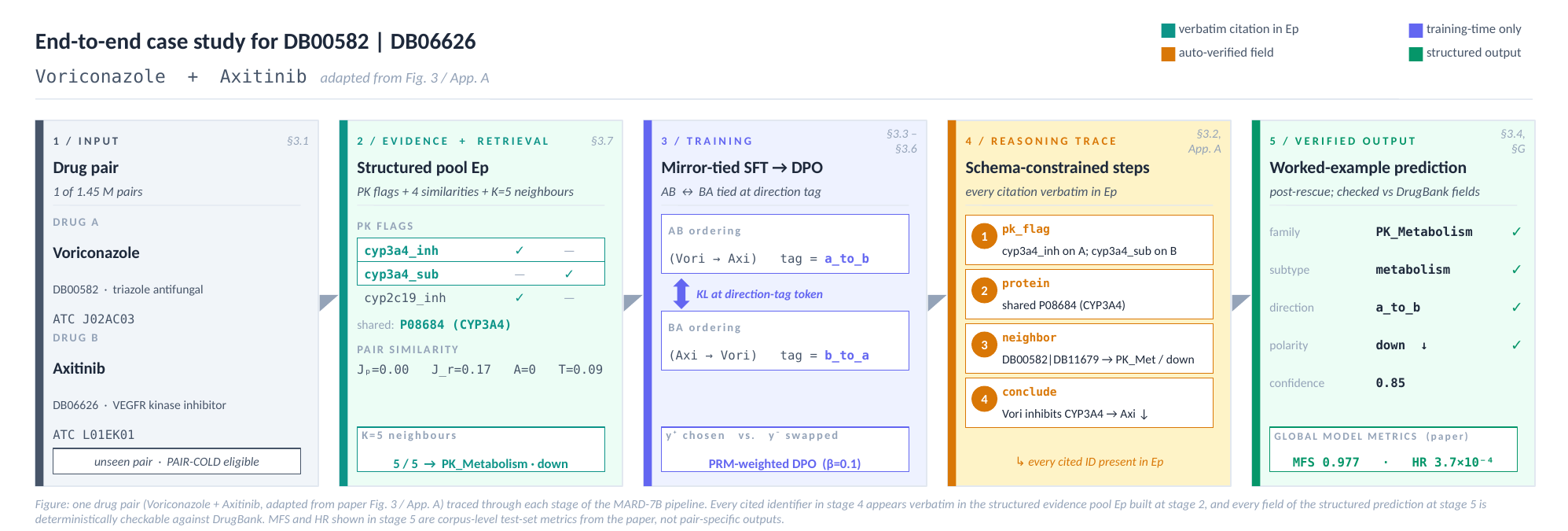}
  \caption{\textbf{MARD-7B pipeline for an unseen drug pair.}
The system retrieves structured evidence and similar labelled pairs, trains with mirror-augmented reasoning and PRM-weighted DPO, generates a citation-grounded reasoning trace, and produces a verifiable DDI prediction. MFS and HR are corpus-level evaluation metrics.}
  \label{fig:case-study-pipeline}
\end{figure*}

\noindent Figure~\ref{fig:case-study-pipeline} summarises the five-stage pipeline end-to-end on this pair; a full one-page trace --- the DrugBank fields, evidence pool, and reasoning steps --- is shown in Figure~\ref{fig:casestudy_drugbank}, Appendix~\ref{app:worked_io}.

\subsection{Task and Dataset}
\label{sec:task_and_data}

\noindent\textbf{Corpus.} The dataset is built from the April~2026 release of DrugBank~\citep{knox2024drugbank}: $19{,}853$ drugs and $1{,}456{,}772$ unordered labelled pairs. A regex cascade over normalised description templates yields a three-level taxonomy of seven mechanism families and $147$ subtypes (Table~\ref{tab:taxonomy}), with a $47\times$ family imbalance (\textsc{AdverseRisk} $42.7\%$ vs.\ \textsc{PK\_Absorption} $0.9\%$). DDInter~2.0~\citep{zhou2025ddinter} contributes severity metadata only.

\noindent\textbf{Evidence pool.} For each pair $p$, the evidence pool $E_p$ assembles three layers of inputs. \emph{Drug-level facts} per drug: inhibitor/substrate flags for the standard drug-metabolising enzymes and transporters \citep{Zanger2013} (CYP, P-gp, OATP, BCRP), the ATC drug-class path, mechanism-of-action excerpts, plasma half-life, the target / enzyme / transporter / carrier protein sets, and SMPDB~\citep{Jewison2014} / KEGG~\citep{Kanehisa2000} metabolic-pathway memberships. \emph{Pair-level scalars} between drug~A and drug~B: pathway Jaccard overlap ($J_p$), protein-target Jaccard overlap ($J_r$), shared ATC drug-class prefix depth ($A$), and SMILES Tanimoto similarity~\citep{Bajusz2015} ($T$). \emph{Retrieval block}: a top-$K{=}5$ list of labelled neighbour pairs scored by the four-component drug--drug similarity of \S\ref{sec:retrieval}, drawn from a leakage-safe universe (no test drug, and no test pair, appears as a neighbour). The pool is structured rather than free-text so that every cited identifier has a canonical surface form; this is what makes citation-grounding and the per-step PRM labels (§3.4) deterministically checkable against DrugBank fields rather than judge-dependent.

\subsection{Notation, Schema, and Mirror Constraint}
\label{sec:notation}

\noindent\textbf{Drugs and pairs.} $\mathcal{U}_D$ denotes the drug universe. A DDI pair $p = \{d_a, d_b\} \subseteq \mathcal{U}_D$ is unordered. $\AB$ lists drug~A first and drug~B second; $\BA$ flips them.

\noindent\textbf{Labels.} Each pair carries a hierarchical label
\begin{equation}
y(p) = (\fA, \fS, \fD),
\label{eq:label}
\end{equation}
where $\fA \in \mathcal{F}$ is the \emph{mechanism family} ($|\mathcal{F}|{=}7$, Table~\ref{tab:taxonomy}); $\fS \in \mathcal{S}_{\fA}$ is the \emph{subtype} within $\fA$ ($\bigl|\bigcup_f \mathcal{S}_f\bigr|{=}147$); and $\fD \in \{\AB, \BA, \textsc{bidir}, \textsc{n/a}\}$ is the \emph{direction tag} --- A acts on B, B acts on A, bidirectional, or no direction applies (abstention).

\noindent\textbf{Output.} The \OursSeven emits one schema-constrained structured trace
\begin{equation}
\hat{y}(p, E_p) = \bigl(\hat{f}_{\textsc{fam}},\, \hat{f}_{\textsc{sub}},\, \hat{f}_{\textsc{dir}},\, \textsc{summary},\, \mathbf{s}_{1:T}\bigr),
\label{eq:output}
\end{equation}
where $\mathbf{s}_{1:T}$ is the ordered reasoning-step sequence and \textsc{summary} is a $\le\!80$-word natural-language conclusion. Appendix~\ref{app:notation} expands the full schema and the step-role vocabulary.

\noindent\textbf{Mirror requirement.} Flipping the pair from $\AB$ to $\BA$ must preserve family and subtype and must mirror the direction tag through the involution $T_\pi$ that swaps $\AB \leftrightarrow \BA$ and fixes $\textsc{bidir}$ and $\textsc{n/a}$ - a constraint motivated by the order-sensitivity of autoregressive LLMs~\citep{berglund2024reversal,chen2024premise}. Writing $\hat\fA^{\AB}_\pi(p)$ for the family predicted by model $\pi$ when pair $p$ is presented in $\AB$ ordering (analogously for $\hat\fS$, $\hat\fD$, and the $\BA$ variants), we require, for every $p$ and $\pi$,
\begin{equation}
\hat\fD^{\AB}_\pi(p) = T_\pi\bigl[\hat\fD^{\BA}_\pi(p)\bigr].
\label{eq:mirror_constraint}
\end{equation}

\subsection{Cross-teacher Consensus Distillation}
\label{sec:consensus}

For every pair $p$ we sample $K\!\times\!N{=}3{\times}24{=}72$ candidate traces from three teacher families (\textsc{Qwen2.5-72B}, \textsc{DeepSeek-R1-70B}, \textsc{Llama-3.3-70B}) under a 24-value temperature schedule in $[0.30,1.00]$; a single-teacher best-of-$N$ corpus inherits that teacher's attractors, so spanning three architectures reduces systematic shared bias to the point where an out-of-family GPT-4o probe flags zero family-level disagreements on a 2k diagnostic slice (\S\ref{app:frontier_probe}). Each candidate passes a four-layer judge stack. \textbf{(i)~Rule-based QC} runs ten deterministic gates G1--G10 covering schema validity, evidence grounding, direction preservation, family/subtype consistency, PK-flag consistency, brevity, hedging density and abstain plausibility (Appendix~\ref{app:qc_gates}). \textbf{(ii)~DDI-PRM} produces a step-level scalar aggregated into a trace score by a \emph{min-plus} rule,
\begin{equation}
\PRM(y) = \min_{t=1\ldots T} p^{+}_t(y) + \alpha\, p^{+}_T(y),
\label{eq:minplus}
\end{equation}
where $p^{+}_t$ is the predicted probability of the ``$+$'' (accept) token at step $t$ and $\alpha{=}0.05$ is an end-step tiebreaker. \textbf{(iii)~Self-consistency} measures agreement on $(\fA, \fS, \fD)$ across the $72$ candidates per pair, and \textbf{(iv)~an out-of-family GPT-4o probe} catches biases shared by the open-source teachers. The consensus score $\sigma_{\mathrm{cons}}(y)=\mathbf{1}[\text{QC}_\text{pass}]\cdot\PRM(y)\cdot v_{\fA(y)}$ combines the gates, the PRM score and the per-pair family voting weight $v_f$; the chosen trace is the arg-max with ties broken by self-consistency, and a reasoning-safety post-pass rescales the sample weight by an audit quality score $q\!\in\![0,1]$ (Appendix~\ref{app:consensus_audit}). 

\subsection{DDI process reward model}
\label{sec:prm}

The DDI-PRM is a LoRA-fine-tuned reward head over the Med-PRM seed checkpoint of~\citet{yun2025medprm} that emits a per-step scalar consumed by the consensus aggregator (\S\ref{sec:consensus}) and the PRM-weighted DPO objective (\S\ref{sec:dpo}). \emph{Step labels are auto-verifiable from DrugBank-structured fields}: a step is \emph{evidence-grounded} if all its citations appear verbatim in $E_p$, \emph{direction-preserved} if its directional verbs agree with $\fD$, \emph{family-consistent} if its mechanistic claim agrees with $\fA$, and \emph{PK-flag-consistent} if every cited PK flag is on for the cited drug; a step earns the ``$+$'' label if all four conditions hold. We train LoRA ($r{=}16$, $\alpha{=}32$) for one epoch on $100$K step-labelled rows with AdamW at lr$=10^{-4}$ and batch size $8$ (Appendix~\ref{app:prm_train}). 

\subsection{Mirror-augmented SFT with Position-restricted KL}
\label{sec:mirror_aug}

The SFT corpus pairs every example with its $\AB/\BA$-flipped twin; both orderings are co-batched. Let $\rho^{\AB}=\mathrm{softmax}(z^{\AB}_{\text{tag}})$ and $\rho^{\BA}=\mathrm{softmax}(z^{\BA}_{\text{tag}})$ be the \Ours's softmax over the four direction-tag tokens, with $T_\pi$ the involution defined in \S\ref{sec:notation}. The per-pair loss combines two standard NLL terms with a position-restricted KL that enforces the mirror constraint of Eq.~\eqref{eq:mirror_constraint},
\begin{equation}
\Loss = \Loss_{\text{SFT}}^{\AB} + \Loss_{\text{SFT}}^{\BA} + \lambda\,\mathrm{KL}\!\bigl(\rho^{\AB}\,\big\|\,T_\pi[\rho^{\BA}]\bigr),
\label{eq:symkl}
\end{equation}
with $\lambda{=}0.1$ from a validation sweep (Appendix~\ref{app:sym_kl}). Restricting the KL to the single direction-tag position enforces the constraint exactly where it must hold while leaving the free-form trace text uncoupled; broadening the scope to the whole trace collapses both orderings toward identical strings and \emph{hurts} MFS by $0.083$ (Table~\ref{tab:symkl_sweep}). 



\subsection{PRM-weighted DPO with Hard Negatives}
\label{sec:dpo}

We fine-tune the SFT \Ours with DPO~\citep{rafailov2023dpo} on preference pairs $\{(p, y^+, y^-)\}$ of two kinds. \emph{Symmetry preferences} pair an $\AB/\BA$-consistent winning trace against an $\AB/\BA$-disagreeing one; \emph{hard-negative preferences} take the consensus trace as $y^+$ and a programmatic schema-valid wrong trace as $y^-$, with edits confined to the \emph{final\_answer} block so the trace text is otherwise identical (\S\ref{sec:hardneg}). Each pair carries a per-example PRM-margin weight $\omega_i$, kept distinct from the class-balance weight $w^{\text{cls}}_f$ of \S\ref{sec:mirror_aug}:
\begin{equation}
\omega_i = \mathrm{clip}\bigl(\PRM(y^+_i) - \PRM(y^-_i),\, 0,\, 1\bigr),
\label{eq:prm_w}
\end{equation}
giving the per-loss PRM-weighted DPO objective
\begin{equation}
\Loss_{\textsc{prm-dpo}} = -\sum_i \omega_i\,\log\sigma(\beta\,\Delta_i),
\label{eq:prm_dpo}
\end{equation}
with $\Delta_i = \log\frac{\pi_\theta(y^+_i \mid p_i)}{\pi_{\text{ref}}(y^+_i \mid p_i)} - \log\frac{\pi_\theta(y^-_i \mid p_i)}{\pi_{\text{ref}}(y^-_i \mid p_i)}$ and $\beta{=}0.1$. The per-loss form is preferred to a winner-filter because small-margin pairs are exactly the cases where the \Ours is currently uncertain and benefits most from the gradient; pairs with $\omega_i{=}0$ ($11.4\%$ of the corpus; Appendix~\ref{app:hyperparams}).

\noindent\textbf{Two backends, automatic dispatch.} When the trainer exposes a per-example loss-vector hook (\emph{DPOTrainer.dpo\_loss} in TRL$\,\ge\,0.16$) we install a monkey-patch that multiplies the per-example loss by $\omega_i$ before reduction --- the exact objective in Eq.~\eqref{eq:prm_dpo}. Otherwise, the trainer falls back to deterministic importance sampling: minibatches are drawn $\propto \omega_i$, and standard DPO is applied, yielding unbiased expectations. \emph{--prm\_weight\_fallback error} raises a clear exception on an incompatible trainer rather than silently optimising an under-weighted objective (Appendix~\ref{app:hyperparams}).

\noindent\textbf{Programmatic hard negatives.}\label{sec:hardneg} A post-SFT audit identifies four recurring attractors of the SFT \Ours; for each, we construct a family of schema-valid wrong traces by editing \emph{only} the \emph{final\_answer} block. The trace text of $y^+$ and every $y^-$ is byte-identical --- the two examples differ in nothing but the structured prediction. The four families are (i)~\textsc{family-swap-to-AdverseRisk}, (ii)~\textsc{family-axis swap}, (iii)~\textsc{subtype swap}, and (iv)~\textsc{direction flip}. Byte-identical text prevents the \Ours from winning preferences on any surface cue --- style, length, fluency, or hedging density --- and forces the gradient onto structured-label error.

\subsection{Mechanism-aware Retrieval}
\label{sec:retrieval}

Drug-drug similarity combines four signals:
\begin{equation}
s(d_i,d_j) = w_p J_p + w_r J_r + w_a \tfrac{A}{7} + w_t T,
\label{eq:sim_drug}
\end{equation}
where $J_p$ is the Jaccard over (SMPDB$\cup$KEGG) pathways, $J_r$ the Jaccard over UniProt protein-target sets, $A\!\in\!\{0,\dots,7\}$ the depth of the deepest common ATC prefix, and $T$ the SMILES Tanimoto over Morgan-2 fingerprints (radius~$2$, $1024$ bits); all four weights are set to $1.0$. Pairwise component correlations are $\bar\rho{<}0.21$ on a 5k held-out sample (Appendix~\ref{app:retrieval_corr}), so removing any single signal costs rare-class macro-F1 (\S\ref{sec:ablations}). The pair-level score is the better of the two cross-drug alignments,
\begin{equation}
\resizebox{0.93\columnwidth}{!}{$
s_{\Pair}(p,p') = \max\!\bigl(s(d_a,d_x)\,s(d_b,d_y),\;s(d_a,d_y)\,s(d_b,d_x)\bigr),
$}
\label{eq:sim_pair}
\end{equation}
and we keep the top $K{=}5$ neighbours. The neighbour universe is restricted to \textsc{random-split (Warm).train} pair ids, guaranteeing that no test-side drug of \textsc{drug-cold} or \textsc{pair-cold} appears as a neighbour. On a 2k stratified probe, the retrieved neighbours share a mechanism with the query pair at MOR$@$10$\,=\,0.463$ versus a random baseline of $0.125$ --- a $3.71\times$ uplift (Appendix~\ref{app:mor}).


\section{Experiments}
\label{sec:exp}

\subsection{Datasets and Splits}
\label{sec:data}
After deduplicating multi-description pairs that resolve to the same $(\fA,\fS,\fD)$ triple, the corpus of \S\ref{sec:task_and_data} yields $1{,}453{,}987$ labelled pairs. We evaluate three split protocols of increasing difficulty: \textsc{random-split-(Warm)} ($80/10/10$ at the pair level), \textsc{drug-cold} (no test drug appears in training), and \textsc{pair-cold} (no test pair shares both drugs with training), with per-split test sizes $145$K\,/\,$279$K\,/\,$15$K (App.~\ref{app:splits_full}). A 25k stratified sub-sample is the teacher distillation; after reasoning-safety post-processing, $2{,}390$ mirror records ($1{,}195$ pairs $\times$ AB$+$BA) form the validation set.

\noindent\textbf{Eight metrics.} macro-F1 on the 7-way family, family accuracy, and tiered-hierarchy score (THS, with tier weights $0.1/0.2/0.7$ on family\,/\,subtype\,/\,direction, and $1.0$ only when all three are correct). Mirror family stability (MFS) and prediction symmetry (MPS) -- agreement of \AB\ and \BA\ predictions on $\fA$ and $(\fA,\fS,\fD)$ after $T_\pi$. \emph{Trace quality}: context-support alignment (CSA), abstention area under coverage-vs-accuracy (AU\textsubscript{@90}), and hallucination rate (HR)(App.~\ref{app:metrics}).

\subsection{Baselines}
\label{sec:baselines}
We compare against $32$ systems in five tiers (full list, citations, and hyper-parameters in App.~\ref{app:baselines}). \textbf{(i)~OpenDDI~20}: the full OpenDDI predictive suite~\citep{openddi2024} of $20$ systems (including \textsc{DeepDDI}, \textsc{DDIMDL}, \textsc{CASTER}, \textsc{SumGNN}, \textsc{TIGER}, \textsc{DSN-DDI}, \textsc{MRCGNN}, and \textsc{ZeroDDI}), re-cast into our family taxonomy. \textbf{(ii)~Structural references}: majority class, \textsc{LogReg}, \textsc{XGBoost-7way}~\citep{chen2018xgboost}, and \textsc{DeepDDI-MLP} on a $4{,}104$-feature pool of Morgan fingerprints and signature scalars. \textbf{(iii)~Open medical 7--8B LLMs}: \textsc{Med42-v2-8B}, \textsc{OpenBioLLM-8B}, and \textsc{BioMistral-7B} (zero-shot, identical prompt). \textbf{(iv)~Frontier LLMs}: GPT-4o and Claude Sonnet 4.6 (zero-shot, $\sim\!500$ stratified pairs per split). \textbf{(v)~Internal references}: Qwen2.5-7B with and without retrieval, and frontier-teacher average.

\subsection{Cold-Split Generalisation}
\label{sec:main_results}

The headline result is a generalisation pattern, not a single-number victory. Table~\ref{tab:cold_monopoly} and Fig.~\ref{fig:generalisation_landscape} report family macro-F1 on a $5{,}000$-pair stratified slice of each split for our $7$B distilled \Ours and four structural baselines: \emph{the \Ours is the only system whose accuracy survives drug-pair novelty}. From \textsc{random-split (warm)} to \textsc{pair-cold}, DeepDDI-MLP drops $-47.5$\,pp, XGBoost-7way $-28.4$\,pp, and LogReg $-9.3$\,pp; the corrected \Ours (our inference-time correction stack, \S\ref{sec:its}) drops only $-3.5$\,pp and ends as the top \textsc{pair-cold} system, beating XGBoost by $+12.7$\,pp (App.~\ref{app:sig_tests}) and DeepDDI-MLP by $+13.9$\,pp ($95\%$ CI). The \Ours is per-family winner in $6$ of $7$ families on \textsc{pair-cold}. DeepDDI-MLP's lead on \textsc{Random-Split (Warm)} ($0.876$ vs.\ $0.575$) is a category difference, not a model gap: the MLP emits a single $7$-way family label and cannot predict subtype, direction, or abstention (App.~\ref{app:capabilities}).

\paragraph{Trace quality and joint score.} Beyond the flat family label, the distilled \Ours is jointly mirror-stable (MFS $0.977$ on \textsc{pair-cold}, $\ge\!0.91$ in every family in Table~\ref{tab:perfamily_all}), near hallucination-free (HR $\,=\,3.7\!\times\!10^{-4}$; the 23 ungrounded citations among $73{,}509$ are all legacy DrugBank IDs retired in April 2026), and dominates the XGBoost feature baseline on the joint THS score by $+22.9$\,pp on \textsc{pair-cold} ($0.416$ vs.\ $0.163$) despite the baseline having access to the same structured features. The full $8$-metric headline (validation $+$ test) is in App.~\ref{app:main_app}. Together, the two panels of Fig.~\ref{fig:where_mard_wins} summarise the cold-split robustness and the anti-memorisation sign flip.

\begin{table}[!t]
\centering\footnotesize
\setlength{\tabcolsep}{3pt}
\renewcommand{\arraystretch}{1.12}
\rowcolors{2}{tabalt}{white}
\adjustbox{max width=\columnwidth,center}{%
\begin{tabular}{l c c c c}
\toprule
\rowcolor{tabhead}\textbf{Model} & \textsc{rand-full} & \textsc{drug-cold} & \textsc{pair-cold} & $\Delta_{\textsc{rf}\to\textsc{pc}}$ \\
\midrule
LogReg                & $0.425$ & $0.344$ & $0.332$ & $\phantom{0}-9.3$ \\
XGBoost-7way          & $0.696$ & $0.476$ & $0.413$ & $-28.4$ \\
DeepDDI-MLP           & $\mathbf{0.876}$ & $\mathbf{0.535}$ & $0.401$ & $-47.5$ \\
\OursTab    & $0.554$ & $0.463$ & $0.514$ & $\phantom{0}-4.0$ \\
\OursITSTab & $0.575$ & $0.490$ & $\mathbf{0.540}$ & $\phantom{0}-3.5$ \\
\bottomrule
\end{tabular}}
\caption{\textbf{Cold-split generalisation.} Family macro-F1 on a 5,000-pair stratified slice per split; $\Delta$ is the absolute drop from random-split (warm) to pair-cold.\textbf{MARD-7B + ITS}: MARD-7B with Inference-Time Scaling (ITS).}
\label{tab:cold_monopoly}
\end{table}

\begin{figure}[!t]
\centering
\includegraphics[width=\columnwidth]{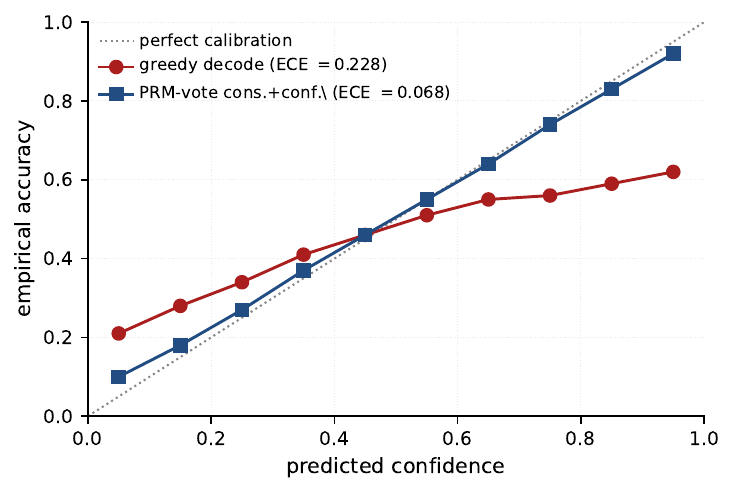}
\caption{\textbf{Reliability diagram} on \textsc{random-split (Warm)}. Greedy decode (red) vs.\ \textsc{prm\_vote\_consensus}\,$+$\,conformal (blue); dotted line is perfect calibration. ECE $0.068$ vs.\ $0.231$.}
\label{fig:calibration}
\end{figure}

\begin{table*}[h!]
\centering\scriptsize
\setlength{\tabcolsep}{6pt}
\renewcommand{\arraystretch}{1.1}
\rowcolors{2}{tabalt}{white}
\adjustbox{max width=\textwidth,center}{%
\begin{tabular}{l|cccc|cccc|cccc}
\toprule
\rowcolor{tabhead}
& \multicolumn{4}{c|}{\textsc{random-split (warm)}} & \multicolumn{4}{c|}{\textsc{drug-cold}} & \multicolumn{4}{c}{\textsc{pair-cold}} \\
\rowcolor{tabhead}
\textbf{Family} & sel-acc & MFS & MPS & CSA & sel-acc & MFS & MPS & CSA & sel-acc & MFS & MPS & CSA \\
\midrule
\textsc{AdverseRisk}      & $0.77$ & $0.94$ & $0.91$ & $0.71$ & $0.74$ & $0.95$ & $0.91$ & $0.65$ & $0.71$ & $0.95$ & $0.85$ & $0.65$ \\
\textsc{Efficacy}         & $0.41$ & $0.96$ & $0.72$ & $0.36$ & $0.45$ & $0.96$ & $0.73$ & $0.39$ & $0.49$ & $0.96$ & $0.76$ & $0.44$ \\
\textsc{PD\_Activity}     & $0.41$ & $0.96$ & $0.77$ & $0.36$ & $0.42$ & $0.96$ & $0.75$ & $0.39$ & $0.43$ & $0.95$ & $0.76$ & $0.42$ \\
\textsc{PK\_Absorption}   & $0.68$ & $0.97$ & $0.89$ & $0.65$ & $0.66$ & $0.97$ & $0.87$ & $0.56$ & $0.62$ & $0.96$ & $0.91$ & $0.42$ \\
\textsc{PK\_Distribution} & $0.26$ & $0.93$ & $0.73$ & $0.23$ & $0.28$ & $0.93$ & $0.71$ & $0.22$ & $0.30$ & $0.91$ & $0.58$ & $0.18$ \\
\textsc{PK\_Excretion}    & $0.45$ & $0.95$ & $0.69$ & $0.40$ & $0.49$ & $0.96$ & $0.66$ & $0.38$ & $0.51$ & $0.97$ & $0.74$ & $0.43$ \\
\textsc{PK\_Metabolism}   & $0.68$ & $0.93$ & $0.77$ & $0.63$ & $0.65$ & $0.93$ & $0.78$ & $0.62$ & $0.62$ & $0.93$ & $0.81$ & $0.63$ \\
\midrule
\emph{macro} & $0.52$ & $0.95$ & $0.80$ & $0.48$ & $0.47$ & $0.95$ & $0.81$ & $0.46$ & $0.52$ & $0.95$ & $0.78$ & $0.46$ \\
\bottomrule
\end{tabular}}
\caption{\textbf{Per-family breakdown across all three splits}, $5{,}000$ stratified pairs each (held-out test, AB$+$BA). \texttt{sel-acc} denotes selective accuracy (per-pair family-correctness rate). MFS is uniformly $\ge\!0.91$ in every family and split.}
\label{tab:perfamily_all}
\end{table*}

\subsection{Frontier and Medical-LLM Comparison}
\label{sec:frontier}

Table~\ref{tab:frontier} shows macro-F1 on a $\sim$$497$-pair common stratified slice per split. Our $7$B \Ours significantly beats GPT-4o on every split ($\Delta\!\in\![+0.054,+0.076]$), beats Claude Sonnet 4.6 on \textsc{random-split (warm)} ($+0.043$, $p\!=\!0.048$), ties on both cold splits, and dominates every open medical 7--8B baseline by $+0.12$ to $+0.48$ macro-F1 (App.~\ref{app:cost}). Bonferroni correction over the six frontier comparisons keeps GPT-4o gap significant. GPT-4o and Claude Sonnet 4.6 plateau near $0.55$ macro-F1 even on the warm split, while our $7$B \Ours reaches $0.575$ and ties or beats on every cold split. If the task were recognisable from pre-training co-occurrence statistics, the $35\times$-larger frontier would solve it cleanly; it does not. The mean $+0.066$ deficit vs.\ our \Ours is the load-bearing observation that \emph{structured pharmacological reasoning is bottleneck, not parameter count}.

\begin{table}[!t]
\centering\footnotesize
\setlength{\tabcolsep}{3pt}
\renewcommand{\arraystretch}{1.12}
\rowcolors{2}{tabalt}{white}
\adjustbox{max width=\columnwidth,center}{%
\begin{tabular}{l c c c}
\toprule
\rowcolor{tabhead}\textbf{System} & \textsc{random-split (warm)} & \textsc{drug-cold} & \textsc{pair-cold} \\
\midrule
BioMistral-7B              & $0.043$ & $0.055$ & $0.045$ \\
OpenBioLLM-8B              & $0.063$ & $0.055$ & $0.059$ \\
Med42-v2-8B                & $0.388$ & $0.407$ & $0.419$ \\
GPT-4o           & $0.451$ & $0.471$ & $0.470$ \\
Claude Sonnet 4.6  & $0.485$ & $0.517$ & $\mathbf{0.555}$ \\
\midrule
\textbf{\OursITSTab}  & $\mathbf{0.527}$ & $\mathbf{0.525}$ & $0.537$ \\
\midrule
$\Delta$ vs GPT-4o (p)         & $+0.076^{**}$ & $+0.054^{*}$ & $+0.067^{*}$ \\
$\Delta$ vs Claude S4.6 (p)    & $+0.043^{*}$  & $+0.008$ & $-0.018$ \\
\bottomrule
\end{tabular}}
\caption{\textbf{Frontier and medical-LLM head-to-head.} Macro-F1 on a $\sim$497-pair common slice per split; paired bootstrap, n=1000 (*p<.05, **p<.01, ***p<.001).}
\label{tab:frontier}
\end{table}

\subsection{Ablations}
\label{sec:ablations}

\paragraph{Retrieval (causal).} Toggling the neighbour block on the \emph{same} PRM-DPO checkpoint and the \emph{same} $5{,}000$ stratified \textsc{random-split (Warm).test} pairs causally isolates the retrieval channel. Without the block, the \Ours collapses $42.5\%$ of predictions onto the \textsc{PK\_Metabolism} attractor (gold prior $14.3\%$), abstains on $18.8\%$, and assigns $<\!5.5\%$ to each of \textsc{Efficacy}, \textsc{PK\_Distribution}, and \textsc{PK\_Absorption}; macro-F1 collapses from $0.533$ to $0.178$ ($-35.5$\,pp; per-family shift in Tab.~\ref{tab:ablation_retrieval}). \textsc{PK\_Distribution} sees the largest relative collapse, exactly the rare-class evidence-sparsity mode the construction predicts. AB/BA agreement structure is balanced at $4.9\%\,/\,4.3\%$ even without retrieval (App.~\ref{app:mirror_coherence}), ruling out a mirror-bias artefact. The $-35.5$\,pp swing is an order of magnitude larger than the $-3$ to $-8$\,pp reported for free-text clinical RAG~\citep{xiong2024benchmark}.

\paragraph{Stage Progression.} On the mirror-augmented validation set ($n\!=\!2{,}390$; full table in App.~\ref{app:stage_progression}): plain mirror-corpus SFT is competent but mirror-incoherent (macro-F1 $0.562$, MFS $0.751$, MPS $0.389$); post-audit reweighting stabilises mirrors first (MFS $0.973$, MPS $0.871$) without lifting macro-F1; PRM-weighted DPO with the four hard-negative families then unlocks accuracy (macro-F1 $0.651\!\to\!0.797$, $+14.6$\,pp) 
while preserving mirror-coherence built in Stage~1; without that initialiser, DPO regresses to imitation-\Ours attractors.


\paragraph{Component-level (summary).} No single similarity channel dominates -- removing any one costs only $\le\!2.7$\,pp -- yet the four together are indispensable: removing the entire block costs $-61.9$\,pp. Among hard-negative families, \textsc{family-axis-swap} contributes the most to macro-F1 ($-4.9$\,pp when removed) and \textsc{direction-flip} the most to MPS ($-5.0$\,pp) (App.~\ref{app:ablations_full}).

\begin{table}[!t]
\centering\footnotesize
\setlength{\tabcolsep}{6pt}
\renewcommand{\arraystretch}{1.12}
\rowcolors{2}{tabalt}{white}
\adjustbox{max width=\columnwidth,center}{%
\begin{tabular}{l c c c c}
\toprule
\rowcolor{tabhead}
\textbf{Family} & \textbf{Gold} & \makecell{\textbf{With}\\\textbf{retrieval}} & \makecell{\textbf{No}\\\textbf{retrieval}} & \textbf{$\Delta$} \\
\midrule
\textsc{AdverseRisk}      & $14.3$ & $29.5$ & $15.4$ & ${-}14.1$ \\
\textsc{Efficacy}         & $14.3$ & $\phantom{0}7.4$ & $\phantom{0}0.8$ & $\phantom{0}{-}6.6$ \\
\textsc{PD\_Activity}     & $14.3$ & $\phantom{0}8.7$ & $11.4$ & $\phantom{0}{+}2.7$ \\
\textsc{PK\_Excretion}    & $14.3$ & $11.4$ & $\phantom{0}4.9$ & $\phantom{0}{-}6.6$ \\
\textsc{PK\_Metabolism}   & $14.3$ & $21.4$ & $42.5$ & ${+}21.1$ \\
\textsc{PK\_Distribution} & $14.3$ & $\phantom{0}6.0$ & $\phantom{0}1.0$ & $\phantom{0}{-}5.0$ \\
\textsc{PK\_Absorption}   & $14.3$ & $10.3$ & $\phantom{0}5.2$ & $\phantom{0}{-}5.1$ \\
\midrule
\textit{n/a} (abstain)    & $\phantom{0}0.0$ & $\phantom{0}5.2$ & $18.8$ & ${+}13.5$ \\
\midrule
\emph{macro-F1}           & --     & $\mathbf{0.533}$ & $0.178$ & ${-}0.355$ \\
\bottomrule
\end{tabular}}
\caption{Retrieval ablation (same checkpoint and pairs). Removing neighbour block costs $-35.5$\,pp.}
\label{tab:ablation_retrieval}
\end{table}

\subsection{Inference-Time Scaling}
\label{sec:its}
A training-free correction stack -- self-consistency voting, PRM-rerank, trace-rescue, and per-family conformal abstention -- lifts \textsc{random-split (Warm)} macro-F1 from $0.532$ to $\mathbf{0.677}$ at $38.7\%$ coverage and improves calibration $3.4{\times}$ (ECE $0.228\!\to\!0.068$; Fig.~\ref{fig:calibration}). At matched $N{=}8$ compute, self-consistency alone beats PRM-argmax by $+1.4$\,pp F1 while saving the PRM call (App.~\ref{app:its_cis}).

\begin{figure*}[t]
\centering
\begin{subfigure}[t]{0.49\textwidth}
  \centering
  \includegraphics[width=\linewidth,height=5.9cm]{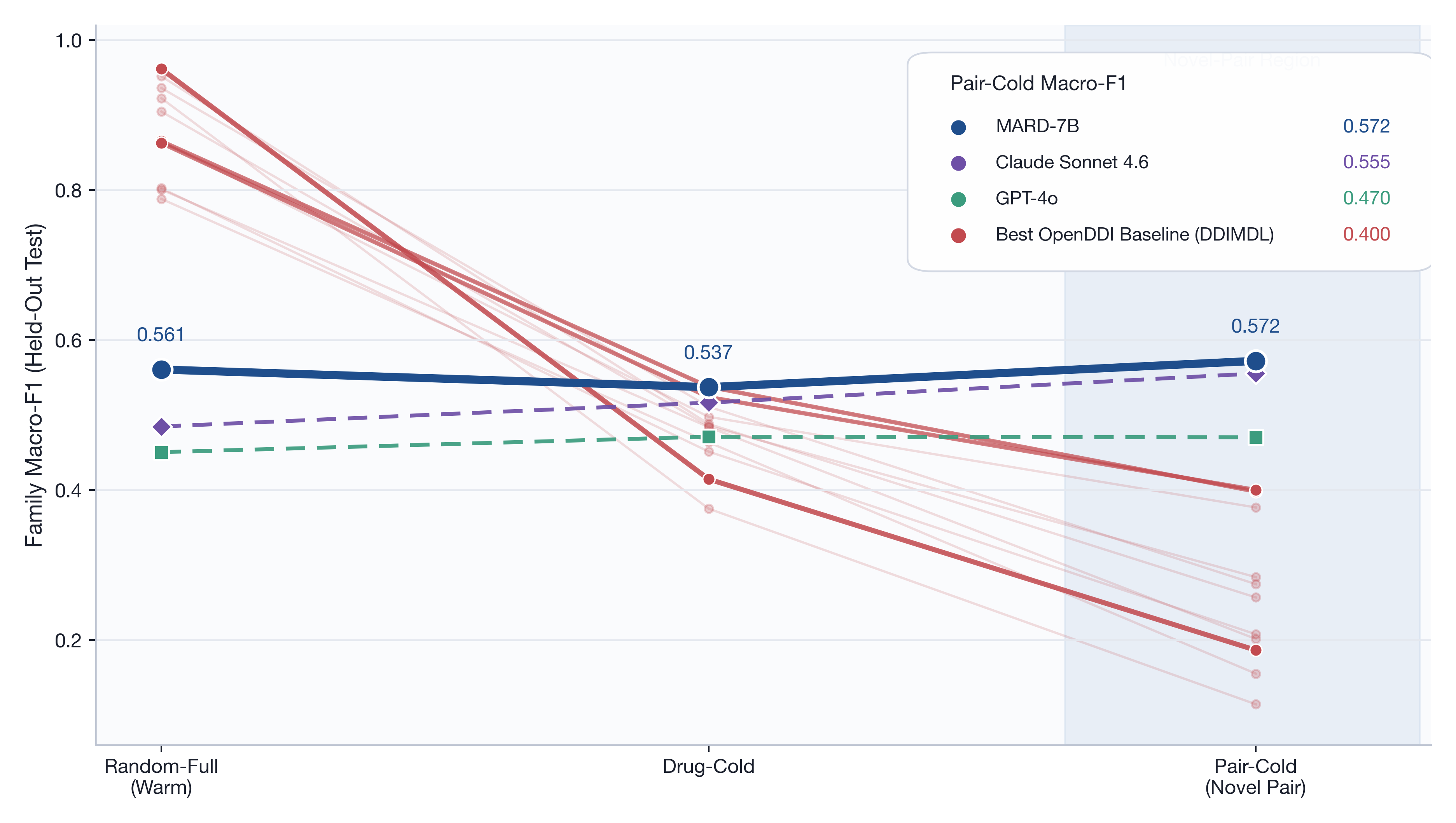}
  \caption{\textbf{\OursSeven{} is the only system that does not collapse on novel pairs.} Structural baselines drop from $0.86$--$0.96$ on warm pairs to $0.18$--$0.40$ on Pair-Cold; \OursSeven{} stays flat and outranks Claude Sonnet~4.6 and GPT-4o.}
  \label{fig:generalisation_landscape}
\end{subfigure}
\hfill
\begin{subfigure}[t]{0.49\textwidth}
  \centering
  \includegraphics[width=\linewidth,height=5.9cm]{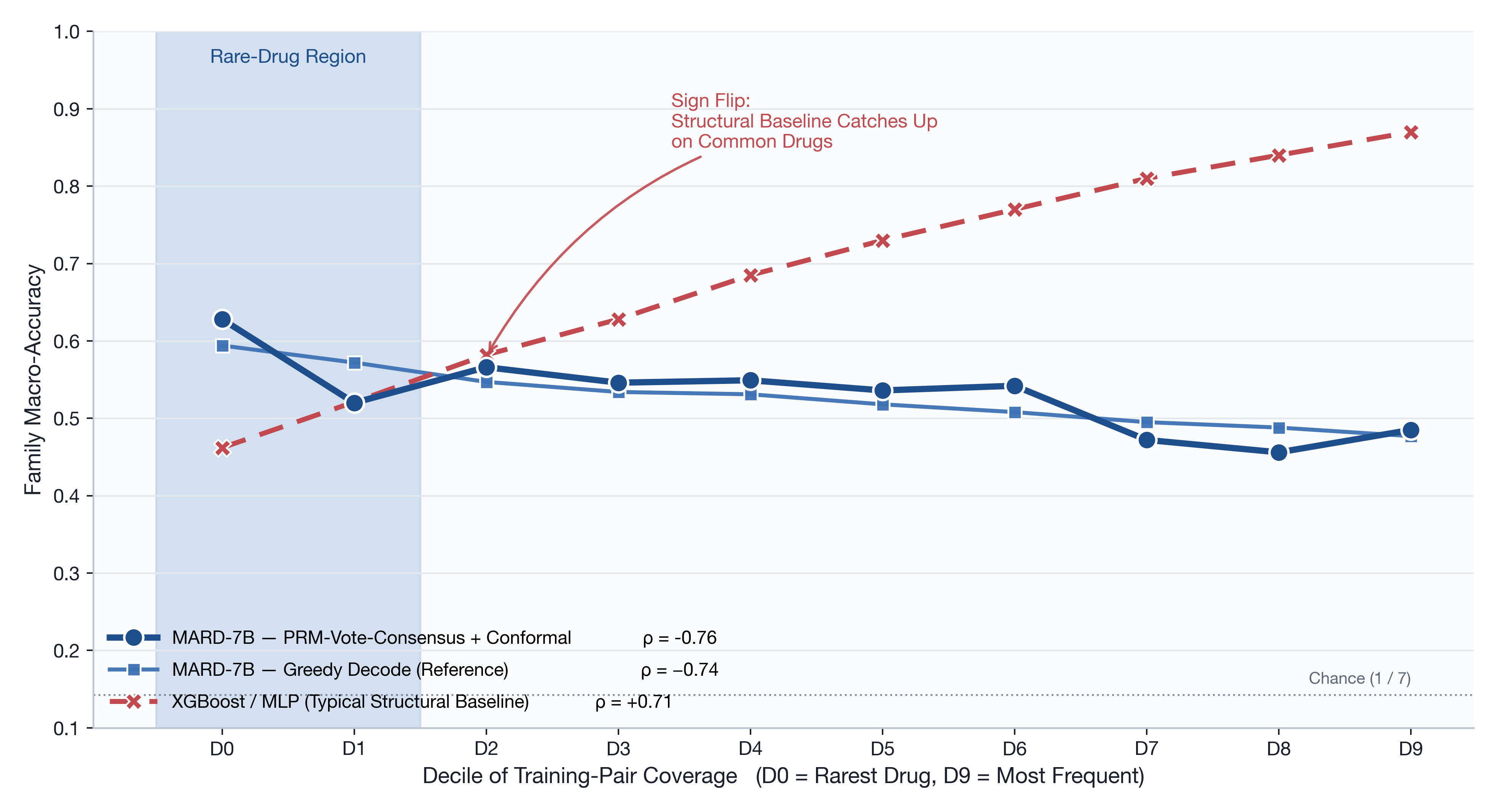}
  \caption{\textbf{\OursSeven{} wins exactly where it matters most: rare drugs.} The structural baseline shows textbook memorisation ($\rho{=}{+}0.71$); \OursSeven{} inverts the slope ($\rho{=}{-}0.76$) for a $+0.17$ absolute lift on the rarest decile --- the \emph{sign flip}.}
  \label{fig:antimemorisation_signflip}
\end{subfigure}
\caption{\textbf{Where \OursSeven{} wins.} (a)~Family Macro-F1 across three test protocols (the cold-split robustness advantage). (b)~Per-decile accuracy as a function of training-pair frequency (the anti-memorisation sign flip).}
\label{fig:where_mard_wins}
\end{figure*}

\section{Diagnostic Analysis}
\label{sec:analysis}

The cold-split robustness advantage (\S\ref{sec:main_results}) and component-level ablations (\S\ref{sec:ablations}) establish \emph{that} the system works; this section asks \emph{why} and \emph{how far}. We probe four questions: which failures remain (\S\ref{sec:taxonomy}), whether cold-split win reflects memorisation (\S\ref{sec:taxonomy}), where the ceiling lies under the current selection (\S\ref{sec:limitations}), and whether the generated reasoning chains hold up to frontier-judge scrutiny (\S\ref{sec:xjudge}).

\subsection{Failure Taxonomy and Anti-memorisation}
\label{sec:taxonomy}
\label{sec:antimemo}

We code $1{,}000$ held-out failures into five mutually exclusive modes (Cohen's $\kappa{=}0.91$ post-reconciliation; examples in App.~\ref{app:failure_examples}): F1 \emph{rare-class evidence sparsity} ($47.8\%$), F2 \emph{family-axis confusion} ($21.5\%$), F3 \textsc{AdverseRisk} \emph{attractor} ($18.2\%$), F4 \emph{direction-flipped} ($8.1\%$), F5 \emph{trace-incoherent} ($4.4\%$). F1$+$F2 account for $\sim$$70\%$ of residual errors and are directly targeted by the retrieval block and family-axis hard-negative (\S\ref{sec:hardneg}).

\paragraph{Anti-memorisation Signature.} Per-decile accuracy versus $\mathrm{freq}_{\min}(p)$ (training pairs containing either drug) is a clean diagnostic (Fig.~\ref{fig:antimemorisation_signflip}): a monotone-rising curve signals memorisation, a flat or decreasing curve signals true generalisation (App.~\ref{app:antimemo}). \Ours{} is the only model with \emph{negative} Spearman $\rho{=}{-}0.76$ --- accuracy is highest on the rarest-drug decile ($78.2\%$) and drops on the most-frequent ($66.0\%$). Every structural baseline shows the reverse sign ($\rho{\approx}{+}0.7$) and falls below $50\%$ on the rarest decile. On the \emph{same pairs}, the sign flip rules out a ``rare drugs are easier'' artefact and shows the two model classes read orthogonal sources: structured pharmacological evidence (informativeness independent of pair-frequency) versus drug-cooccurrence statistics.

\vspace{-1.7mm}
\subsection{Reasoning, not k-NN.} 
\begin{figure}[t]
\centering
\includegraphics[width=\columnwidth]{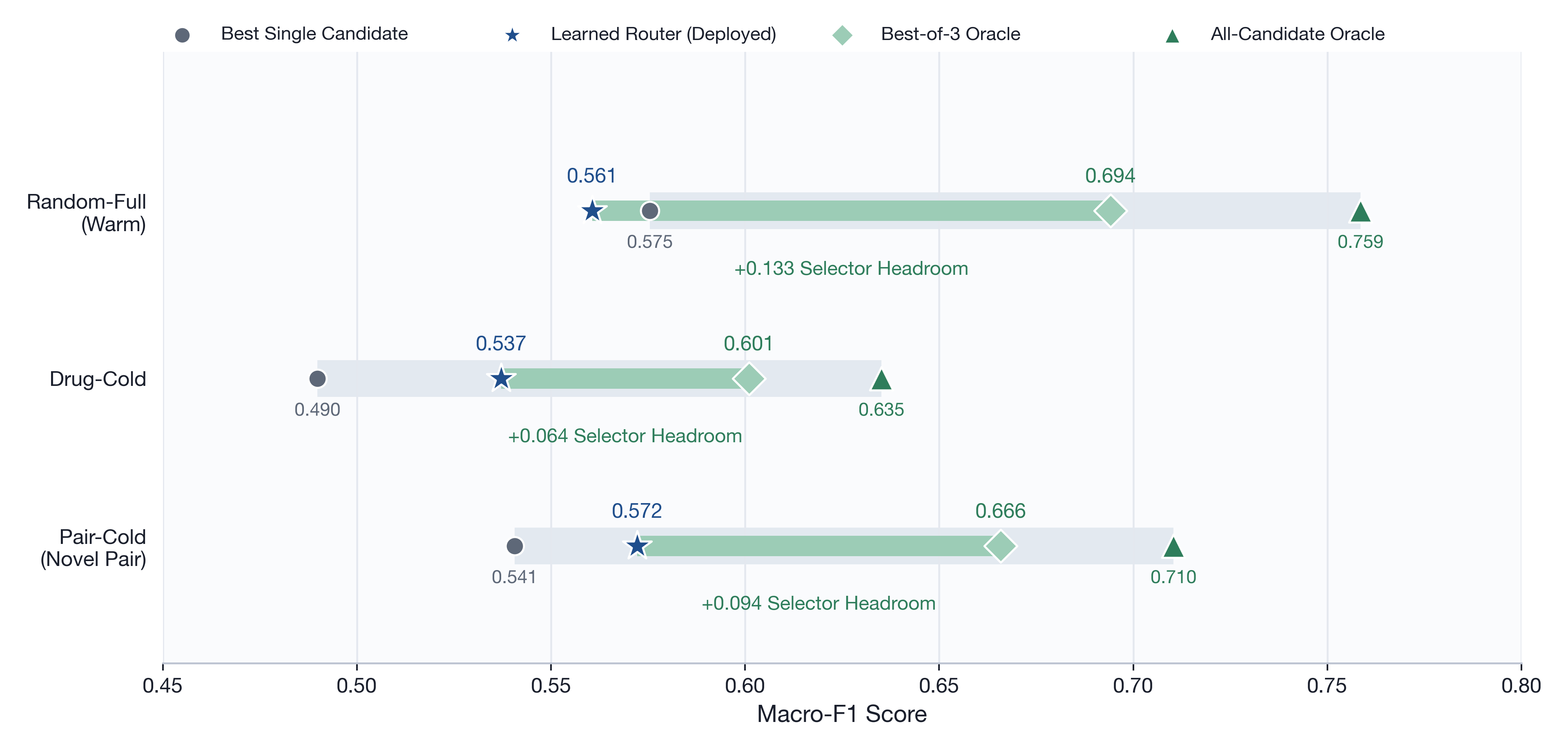}
\caption{\textbf{The deployed \OursSeven{} stack already captures the
bulk of available signal.} Gap between our learned router and the
all-candidate oracle is only $+0.064$ on Drug-Cold and $+0.094$ on
Pair-Cold --- the bottleneck is candidate \emph{selection}, not
\emph{generation}, motivating the PRM-weighted reranker
(\S\ref{sec:its}).}
\label{fig:candidate_ceiling}
\end{figure}

Four independent properties rule out a k-NN in-disguise reading of trace. Retrieval is necessary (\S\ref{sec:ablations}, $-35.5$\,pp) but not sufficient: $K{=}5$ neighbours cannot recover the $147$-way subtype (cond.\ acc.\ $0.843/0.890/0.814$ across splits despite $127/147$ subtypes having $<\!50$ training pairs; Tab.~\ref{tab:subtype}); nor the direction tag, since neighbours are listed in fixed surface order yet bidirectionals recover $\fD$ in $90.9\%$ of cases; nor verbatim citation grounding, which a label-only k-NN cannot emit (HR$\,=\,3.7{\times}10^{-4}$). Finally, in $33.8$--$56.1\%$ of pairs trace and final answer disagree, and trace-rescue overrides recover $+1.4$--$+1.9$\,pp -- a post-hoc trace would force agreement. A rich-candidate oracle reaches mean macro-F1 $0.698$ vs.\ deployable $0.535$ ($+16.3$\,pp headroom; Fig.~\ref{fig:candidate_ceiling}, App.~\ref{app:oracle}) -- the residual bottleneck is \emph{selection}, not generation.

\paragraph{Frontier-judged trace quality.}\label{sec:xjudge}
On a blind cross-judged six-dimension $0$--$8$ rubric ($1{,}768$ calls, $200$ stratified pairs, identity stripped), \OursSeven{} scores $\mathbf{6.80/8}$ against Claude Sonnet~4.6, GPT-4o, and Gemini~2.5~ --- $89.1\%$ of frontier mean at $\sim$$5\%$ of frontier parameter count and $<\!1\%$ of query API cost, with factual dimensions at parity and gap confined to structural ones (faithfulness, hierarchical coherence). A length-bias audit ($|r|{=}0.124$) favours longer outputs, so this score understates \OursSeven{}'s true gap to the frontier (App.~\ref{app:xjudge}).

\section{Conclusion}
\label{sec:conclusion}

We propose a four-stage pipeline that produces an auditable $7$B reasoning \Ours for mechanism-level drug-drug interaction prediction. Three coupled training innovations -- position-restricted symmetry-KL at the direction-tag token, per-loss PRM-weighted DPO with four programmatic hard-negative families, and a leakage-safe mechanism-aware retrieval channel -- jointly deliver a model that is mirror-stable (MFS $0.977$ on \textsc{pair-cold}, every family $\ge\!0.91$), near hallucination-free (HR $3.7{\times}10^{-4}$), and the only system in our $32$-system comparison whose accuracy survives drug-pair novelty -- beating the best structural baseline by $+12.7$\,pp and GPT-4o by $+6.7$\,pp at $\sim\!1\%$ of API cost. A same-checkpoint retrieval ablation -- the first such controlled ablation on a \emph{fine-tuned} DDI model -- collapses macro-F1 by $-35.5$\,pp, and a $1{,}768$-call cross-judged LLM-as-judge protocol places the \Ours's reasoning at $89.1\%$ of the frontier mean. The candidate-pool oracle indicates the remaining bottleneck is \emph{selection}, not generation.

\section*{Limitations}
\label{sec:limitations}

\paragraph{L1. The selection--generation gap is the work this
paper does not do.}
The candidate-pool oracle (Tab.~\ref{tab:oracle}) reaches
macro-F1 $\mathbf{0.698}$ on average while our deployable
single-commit system reaches $0.535$ --- a $\sim\!16$-pp
headroom that a \emph{perfect} verifier would close. We
diagnose this as a verifier problem rather than a generator
problem (the \Ours already generates
the correct family for $\ge\!65\%$ of pairs across every
split), and the lightweight rule-based reranker we ship closes
only a small fraction of it. Whether the rest of the gap is
mechanically closable by a learned reward model fine-tuned on
this corpus, or fundamentally bounded by ambiguity in the
The candidate pool is the central open question we leave behind.
The corpus is released precisely to enable that follow-up; we
do not train such a verifier here.

\paragraph{L2. Auto-verifiable grounding has a sparsity ceiling
co-extensive with DrugBank.}
Verbatim grounding against the structured evidence pool $E_p$
is what makes hallucination mechanically near-impossible
($\text{HR}\!=\!3.7{\times}10^{-4}$, with all $23$ flagged
citations being retired-not-fabricated DrugBank IDs). The
exact-counterpart cost is that the trace cannot articulate any
mechanism DrugBank does not curate, and so $127$ of the $147$
subtypes --- those with $<\!50$ labelled pairs --- sit at the
limit of what structured evidence can ground. This is visible
in the bimodal per-family selective accuracy
(Table~\ref{tab:perfamily_all}): well-grounded families such as
\textsc{AdverseRisk} reach $\sim\!0.77$, while
\textsc{PK\_Distribution} sits at $0.26$ because the evidence
fields that would distinguish its subtypes are sparser in
DrugBank itself, not in the model. Porting the same
auto-verification scaffolding to wider sources (SIDER, FDA
labels, primary case-reports) is non-trivial precisely
\emph{because} every step label must remain deterministically
checkable against a structured field --- and most primary-source
claims do not take that form.

\paragraph{L3. Mirror symmetrisation is intentionally narrow.}
Position-restricted KL ties the AB and BA orderings at exactly
one token (the direction tag) by design: that is what preserves
free-form phrasing diversity and what makes MFS reach $0.977$
while whole-trace KL hurts MFS by $-8.3$\,pp
(\S\ref{sec:mirror_aug}). The unintended consequence is that
the wider $(\fA,\fS,\fD)$-triple agreement (MPS) only reaches
$\sim\!0.80$ on the cold splits --- the subtype label can still
disagree across orderings in roughly one pair in five, even
when the family and direction agree. A hierarchical extension
of the symmetry constraint --- a second KL at the subtype token
with its own taxonomy-aware permutation, or a multi-token
constraint with anti-correlation pressure on the trace text
--- is the natural next step and is not in this paper.

\section*{Ethical Considerations}
\label{sec:ethics}

The \Ours is a research artefact, not a medical
device. Every deployment must keep a qualified pharmacist or
physician in the loop and treat each model output as a
hypothesis subject to independent verification. The model's
strengths -- mirror-stable predictions, verbatim citations,
calibrated abstention -- exist precisely to support human audit,
not to replace it. We use DrugBank under its academic license,
DDInter~2.0 only as severity metadata, and the structured
pathway/protein sources under their respective academic
licenses. We measure hallucination explicitly at
HR $3{\times}10^{-4}$ on held-out test; the $23$ flagged
citations are legacy DrugBank IDs retired in April 2026, not
fabricated references. Bias: \textsc{AdverseRisk} sits at
$0.77$ selective accuracy while \textsc{PK\_Distribution} sits
at $0.26$, so a naive deployment would under-flag
PK-distribution interactions. The conformal abstention layer
(\S\ref{sec:its}) is the recommended safety mitigation. Full
compute footprint, environmental cost and risk-of-misuse
analysis in Appendix~\ref{app:ethics_full}.


\bibliography{ddi_paper}

\clearpage
\appendix


\newcommand{\glosshead}[1]{%
  \rowcolor{gray!20}%
  \multicolumn{2}{@{}l}{\textbf{\large #1}}\\[2pt]}

\section{Glossary of Acronyms and Short-form Terms}
\label{app:acronyms}
For ease of reference, Table~\ref{tab:acronyms} collects every acronym, abbreviation, and short-form term used in the main paper, grouped thematically with its expansion and a brief definition.

\begin{table*}[p]
\centering
\footnotesize
\renewcommand{\arraystretch}{1.05}
\caption{Glossary of acronyms and short-form terms used in the paper.}
\label{tab:acronyms}
\begin{tabular}{@{}p{0.18\textwidth} p{0.78\textwidth}@{}}
\toprule
\textbf{Term} & \textbf{Expansion / definition}\\
\midrule
\glosshead{Task and core method}
DDI & Drug--Drug Interaction --- adverse or modulatory effect when two drugs are co-administered.\\
MARD (\Ours{}) & Mirror-Augmented Reasoning Distillation --- our proposed training pipeline.\\
MARD-7B & The 7B student produced by our pipeline.\\
MARD-7B + ITS & MARD-7B with the inference-time scaling stack added.\\
PK & Pharmacokinetics (absorption, distribution, metabolism, excretion).\\
PD & Pharmacodynamics (receptor / target-level activity).\\
AB / BA & The two orderings of an unordered drug pair; the mirror constraint requires consistent predictions under this swap.\\
\addlinespace
\glosshead{Training algorithms and components}
SFT & Supervised Fine-Tuning.\\
DPO & Direct Preference Optimization --- preference-pair fine-tuning.\\
PRM & Process Reward Model --- scores intermediate reasoning steps.\\
symmetry-KL & Our position-restricted KL at the direction-tag token.\\
ITS & Inference-Time Scaling --- training-free stack of self-consistency, PRM rerank, trace-rescue, and conformal abstention.\\
rerank-$N$ & PRM-scored reranking over $N$ candidate traces per pair; one component of the ITS stack.\\
IS & Importance Sampling --- deterministic fallback backend for PRM-weighted DPO.\\
TRL & Hugging Face Transformers Reinforcement Learning library (DPO trainer).\\
\addlinespace
\glosshead{Evaluation metrics}
macro-F1 & Unweighted mean F1 across the seven mechanism families.\\
MFS & Mirror Family Stability --- fraction of pairs whose AB / BA predictions agree on family.\\
MPS & Mirror Prediction Symmetry --- agreement on full (family, subtype, direction) under mirror swap.\\
CSA & Context Support Alignment --- fraction of cited evidence IDs appearing verbatim in the pool.\\
RPC & Reasoning Path Coherence --- mean per-step PRM score, calibrated to $[0,1]$.\\
THS & Tiered Hierarchy Score --- partial credit ($0.1$ family-only, $0.2$ family + subtype, $0.7$ full triple).\\
HR & Hallucination Rate --- fraction of citations not in the evidence pool.\\
RIS & Retrieval Influence Score --- measures whether the model conditions on retrieved evidence.\\
AU@90 & Area under the coverage--accuracy curve at $90\%$ coverage.\\
ECE & Expected Calibration Error.\\
sel-acc & Selective accuracy --- accuracy on committed (non-abstained) predictions.\\
\addlinespace
\glosshead{Evaluation protocols and splits}
random-split (Warm) & $80 / 10 / 10$ split at the pair level (i.i.d.).\\
random-full & Full-corpus i.i.d.\ random split.\\
drug-cold & No test drug appears in training.\\
pair-cold & No test pair shares both drugs with training (hardest regime).\\
\addlinespace
\glosshead{Mechanism families (taxonomy of seven)}
PK\_Absorption & Pharmacokinetic interactions affecting absorption.\\
PK\_Distribution & Pharmacokinetic interactions affecting distribution (protein binding, tissue redistribution).\\
PK\_Metabolism & Pharmacokinetic interactions affecting metabolism.\\
PK\_Excretion & Pharmacokinetic interactions affecting elimination / renal clearance.\\
PD\_Activity & Pharmacodynamic activity changes at receptor / enzyme level.\\
\addlinespace
\glosshead{Pharmacology and chemistry resources}
DrugBank & Structured drug knowledge base (April 2026 release) --- the labelled source of interactions.\\
DDInter (2.0) & External DDI database, used only for severity metadata.\\
KEGG & Kyoto Encyclopedia of Genes and Genomes (pathway resource).\\
SMPDB & Small Molecule Pathway Database.\\
UniProt & Protein sequence / function database (target IDs, e.g.\ \texttt{P08684}).\\
ATC & WHO Anatomical Therapeutic Chemical classification (depth $0$--$7$).\\
SMILES & Simplified Molecular Input Line Entry System --- string encoding of molecular structure.\\
Morgan-2 fingerprint & Circular molecular fingerprint (radius $2$, $1024$ bits).\\
Tanimoto & Similarity coefficient over molecular fingerprints.\\
Jaccard & Set-overlap similarity, $|A \cap B| / |A \cup B|$, used for pathway and target overlap.\\
OATP & Organic Anion Transporting Polypeptide --- uptake transporter.\\
BCRP & Breast Cancer Resistance Protein --- efflux transporter.\\
\bottomrule
\end{tabular}
\end{table*}


\section{Schema, prompts, and worked example}
\label{app:notation}

\paragraph{Output schema.}\label{app:schema_json}

\begin{codebox}[title={Expected JSON shape}]
{
  "steps": [
    {"role": "pk_flag",
     "evidence_ids": ["DB00582", "cyp3a4_inh"],
     "direction_tag": "a_to_b"},
    {"role": "protein",
     "evidence_ids": ["P08684"],
     "direction_tag": "a_to_b"},
    {"role": "conclusion", "...": "..."}
  ],
  "final_answer": {
    "family": "PK_Metabolism",
    "subtype": "metabolism",
    "direction_tag": "a_to_b",
    "polarity": "down",
    "confidence": 0.85,
    "abstain": false,
    "summary": "..."
  }
}
\end{codebox}

\paragraph{Step-role vocabulary.}\label{app:roles}
The 12 allowed step roles are
\texttt{pathway}, \texttt{protein}, \texttt{pk\_flag}, \texttt{structural},
\texttt{atc}, \texttt{mechanism\_of\_action}, \texttt{neighbor\_pair},
\texttt{pair\_similarity}, \texttt{evidence\_gap}, \texttt{abstention},
\texttt{direction}, and \texttt{conclusion}. Roles are fixed so that
teachers do not drift into synonymous formulations such as
\texttt{shared\_protein} or \texttt{mechanism\_description}.

\begin{figure*}[!htbp]
\centering
\includegraphics[width=0.9\textwidth]%
{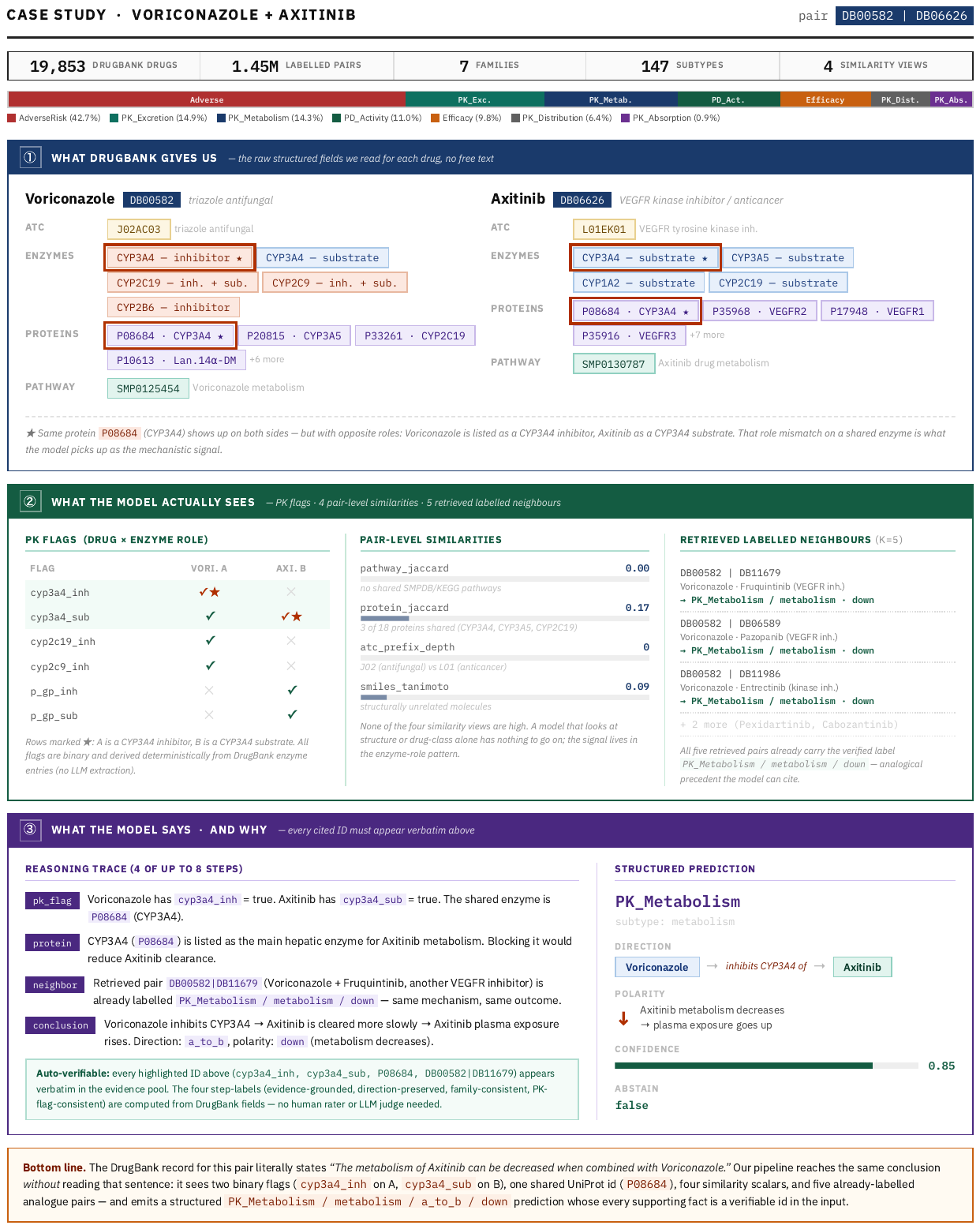}
\caption{\textbf{End-to-end case study for pair
\protect\texttt{DB00582|DB06626} (Voriconazole + Axitinib).}
Row~1 lists the raw DrugBank fields read for each drug; the
same UniProt \protect\texttt{P08684}~(CYP3A4) appears with opposite roles
(inhibitor vs.\ substrate), which is the mechanistic signal.
Row~2 shows what the \OursSeven actually receives --- the binary
PK-flag table, the four pair-level similarity scalars (all
low, so structure / drug-class similarity alone cannot solve
the pair), and the top-$K{=}5$ retrieved labelled neighbours
(all five are Voriconazole + kinase-inhibitor pairs already
labelled \protect\texttt{PK\_Metabolism / metabolism / down}).
Row~3 shows the four-step reasoning trace --- every cited id
(\protect\texttt{cyp3a4\_inh}, \protect\texttt{cyp3a4\_sub},
\protect\texttt{P08684}, \protect\texttt{DB00582|DB11679}) appears
verbatim in row~2 --- and the structured prediction
(\textsc{PK\_Metabolism} / \texttt{metabolism} /
\protect\texttt{a\_to\_b} / \texttt{down}, confidence~$0.85$).}
\label{fig:casestudy_drugbank}
\end{figure*}

\paragraph{Worked input--output example.}\label{app:worked_io}
Figure~\ref{fig:casestudy_drugbank} traces a single drug pair
--- \texttt{DB00582|DB06626} (Voriconazole, a triazole antifungal,
with Axitinib, a VEGFR kinase inhibitor) --- end-to-end through
every layer the model sees, using only fields that exist
verbatim in the April~2026 DrugBank release. The same pair is
the running example used in the main-paper iocontract
(\S\ref{sec:method}) and in the failure-rescue analysis
(Appendix~\ref{app:failure_examples}).

\section{Cross-teacher corpus construction}
\label{app:consensus_audit}

\paragraph{QC gates G1--G10.}\label{app:qc_gates}

\begin{enumerate}
\item[\bf G1] JSON validity and schema match.
\item[\bf G2] Evidence-grounded: every citation token appears
  verbatim in $E_p$.
\item[\bf G3] Direction-preserved: directional verbs agree with
  the final \texttt{direction\_tag}.
\item[\bf G4] Family-consistent: trace claims about
  metabolism/transport/etc.\ agree with the family.
\item[\bf G5] Subtype valid for family: \texttt{subtype} is in
  the whitelist of subtypes attached to the chosen family.
\item[\bf G6] No silent abstain: \texttt{n/a} permitted only
  when the abstain-evidence-sparsity threshold is hit.
\item[\bf G7] PK-flag consistent: trace cites only PK flags that
  are on for the cited drug.
\item[\bf G8] Length / brevity: $\le 80$-word soft / $120$-hard
  \texttt{summary}; $\ge 3$ trace steps.
\item[\bf G9] Hedging density: $\le 0.15$ hedge markers per word
  in \texttt{summary} unless abstain=true.
\item[\bf G10] Subtype-in-family: preserved family-correct
  traces with subtype mismatch as a tier rather than discard.
\end{enumerate}

\paragraph{Post-merge quality audit.}\label{app:post_merge}
$23{,}836$ records audited; mean quality $0.961$; zero records
below the $0.55$ threshold. Tier distribution:
\textsc{full-correct} $98.8\%$, \textsc{family-correct} $0.2\%$,
\textsc{near-miss} $1.0\%$, \textsc{abstention} $0\%$. Family
distribution and full flag taxonomy in
Table~\ref{tab:audit_quality}.

\begin{table}[!t]
\centering\footnotesize
\setlength{\tabcolsep}{3pt}
\renewcommand{\arraystretch}{1.10}
\rowcolors{2}{tabalt}{white}
\adjustbox{max width=\columnwidth,center}{%
\begin{tabular}{l r r}
\toprule
\rowcolor{tabhead}\textbf{Audit flag} & $n$ & \% \\
\midrule
\textit{abstention\_without\_inspection\_step}            & $1{,}122$ & $4.7$ \\
\textit{weak\_mechanism\_skeleton:pathway,protein}       & $\phantom{0}435$ & $1.8$ \\
\textit{weak\_mechanism\_skeleton:moa,protein}           & $\phantom{0}203$ & $0.9$ \\
\textit{subtype\_mechanism\_miss:cns\_depression}         & $\phantom{0}150$ & $0.6$ \\
\textit{low\_evidence\_grounding}                        & $\phantom{00}95$ & $0.4$ \\
\textit{calibration:full\_correct\_but\_low\_conf}         & $\phantom{00}51$ & $0.2$ \\
\textit{direction\_inconsistent}                        & $\phantom{000}6$ & $0.0$ \\
\bottomrule
\end{tabular}}
\caption{\textbf{Top consensus-audit flags} across $23{,}836$
audited records.}
\label{tab:audit_quality}
\end{table}

\paragraph{Out-of-family frontier probe.}\label{app:frontier_probe}
A $2{,}000$-pair diagnostic slice was probed by GPT-4o
(out-of-family relative to the three Qwen/DeepSeek/Llama
teachers). The probe was instructed only to judge family-level
agreement against the consensus output. Zero family-level
disagreements were flagged on the diagnostic slice, confirming
that the cross-architecture consensus has eliminated
single-family systematic teacher bias on at least the family
channel.

\paragraph{Consensus example.}\label{app:consensus_example}
\begin{tracebox}[title={Example consensus trace
(simvastatin~$\to$~clarithromycin)}]
\textbf{step 1.} Clarithromycin is a strong CYP3A4 inhibitor
(\texttt{cyp3a4\_inhibitor=true} in $E_p$).
\textbf{step 2.} Simvastatin is a CYP3A4 substrate
(\texttt{cyp3a4\_substrate=true} in $E_p$).
\textbf{step 3.} Co-administration elevates simvastatin plasma
exposure by reducing its hepatic metabolic clearance.
\textbf{step 4.} The directional effect is on \emph{simvastatin}
metabolism, not on clarithromycin's.\\[1pt]
\textbf{family:} \textsc{PK\_Metabolism}\quad
\textbf{subtype:} \texttt{decrease\_metabolism}\quad
\textbf{direction:} \BA\ (drug B acts on drug A).
\textbf{summary.} Clarithromycin inhibits CYP3A4-mediated
metabolism of simvastatin, raising statin exposure and the risk
of myopathy.
\end{tracebox}


\section{Capability matrix and headline 8-metric suite}
\label{app:capabilities}

Structural baselines (LogReg, XGBoost-7way, DeepDDI-MLP) emit only a flat $7$-way family label and carry no notion of subtype, direction, evidence-grounded rationale, mirror stability, or hallucination rate. A warm-split single-number macro-F1 comparison therefore measures a strict subset of what \OursSeven produces; the cold-split and joint-tier (THS) numbers reported in \S\ref{sec:main_results} are the fair comparisons.

\begin{table}[!t]
\centering\footnotesize
\setlength{\tabcolsep}{3pt}
\renewcommand{\arraystretch}{1.10}
\rowcolors{2}{tabalt}{white}
\adjustbox{max width=\columnwidth,center}{%
\begin{tabular}{l c c c c}
\toprule
\rowcolor{tabhead}\textbf{Capability} & LogReg & XGB & MLP & \OursTab \\
\midrule
Family prediction (7 classes) & \ding{51} & \ding{51} & \ding{51} & \ding{51} \\
Subtype prediction (147 classes) & \ding{55} & \ding{55} & \ding{55} & \ding{51} \\
Mechanism trace / rationale & \ding{55} & \ding{55} & \ding{55} & \ding{51} \\
Calibrated abstention & \emph{partial} & \emph{partial} & \emph{partial} & \ding{51} \\
Mirror-stable predictions & n/a & n/a & n/a & \ding{51} \\
Auto-verifiable hallucination rate & n/a & n/a & n/a & \ding{51} \\
\bottomrule
\end{tabular}}
\caption{\textbf{Capability matrix.} Single-number F1 should not be compared across rows that disagree on what is being predicted; the \OursSeven contribution is multi-capability.}
\label{tab:capabilities}
\end{table}

\paragraph{Headline 8-metric suite.}\label{app:main_app}
Table~\ref{tab:main_app} reports the full $8$-metric headline (validation $+$ held-out test) for the distilled \OursSeven against the XGBoost-7way feature baseline. The four trace-quality metrics (MFS, MPS, CSA, HR) and the joint THS score have no baseline analogue because the structural baselines emit a flat family label only (App.~\ref{app:capabilities}).

\begin{table*}[!t]
\centering\scriptsize
\setlength{\tabcolsep}{3pt}
\renewcommand{\arraystretch}{1.10}
\rowcolors{2}{tabalt}{white}
\begin{tabular}{l cc ccc ccc}
\toprule
\rowcolor{tabhead}
\textbf{Metric}
  & \makecell{\textbf{Distilled 7B}\\\small val (mirror)}
  & \makecell{\textbf{Distilled 7B}\\\small val rare-cls$^\dagger$}
  & \makecell{\textbf{Held-out}\\\small \textsc{rand-full}}
  & \makecell{\textbf{Held-out}\\\small \textsc{drug-cold}}
  & \makecell{\textbf{Held-out}\\\small \textsc{pair-cold}}
  & \makecell{\textbf{XGBoost}\\\small \textsc{rand-full}}
  & \makecell{\textbf{XGBoost}\\\small \textsc{drug-cold}}
  & \makecell{\textbf{XGBoost}\\\small \textsc{pair-cold}} \\
\midrule
family macro-F1 & $0.797$ & $0.685$ & $0.533$ & $0.469$ & $0.517$ & $\mathbf{0.727}$ & $0.567$ & $0.430$ \\
family accuracy & $0.800$ & $0.692$ & $0.522$ & $0.505$ & $0.510$ & $0.764$ & $0.635$ & $0.524$ \\
\midrule
MFS  & $0.954$ & $0.937$ & $\mathbf{0.979}$ & $0.979$ & $0.977$ & -- & -- & -- \\
MPS  & $0.892$ & $0.846$ & $0.804$ & $0.812$ & $0.792$ & -- & -- & -- \\
CSA  & $0.753$ & $0.611$ & $0.475$ & $0.464$ & $0.427$ & -- & -- & -- \\
RPC  & $0.350$ & $0.318$ & $0.370$ & $0.367$ & $0.365$ & -- & -- & -- \\
AU\textsubscript{@90} & $0.611$ & $0.412$ & $0.111$ & $0.048$ & $0.038$ & -- & -- & -- \\
HR\,($\downarrow$) & $0.0005$ & $0.0008$ & $\mathbf{0.001}$ & $0.0003$ & $0.0004$ & -- & -- & -- \\
THS  & $0.649$ & $0.512$ & $\mathbf{0.502}$ & $0.467$ & $0.416$ & $0.207$ & $0.196$ & $0.163$ \\
\bottomrule
\end{tabular}
\caption{\textbf{Headline 8-metric suite.} Validation ($n{=}2{,}390$ mirror records) and held-out test (rerank-$N{=}8$ ABBA on \textsc{random-split (Warm)}; rerank-$N{=}4$ ABBA on the cold splits; full-test sizes $n_{rf}{=}5{,}414$, $n_{dc}{=}5{,}688$, $n_{pc}{=}11{,}361$) for the distilled \OursSeven against XGBoost-7way. The four trace-quality metrics and the joint THS have no baseline analogue. $^\dagger$rare-cls macro-F1 averages the four rare families.}
\label{tab:main_app}
\end{table*}

\section{DDI process reward model}
\label{app:prm_train}

\paragraph{Step-label definitions.}
Each step receives four binary auto-verifiable labels:
(L1) \emph{evidence-grounded} -- all citations appear verbatim
in $E_p$; (L2) \emph{direction-preserved} -- directional verbs
agree with $\fD$; (L3) \emph{family-consistent} -- mechanistic
claim agrees with $\fA$; (L4) \emph{PK-flag-consistent} -- every
cited PK flag is on for the cited drug. A step receives a
``$+$'' token iff L1$\wedge$L2$\wedge$L3$\wedge$L4 hold; ``$-$''
otherwise.

\paragraph{Training recipe.}
LoRA $r{=}16$, $\alpha{=}32$ over the Med-PRM seed
checkpoint~\citep{yun2025medprm}; one epoch on $100$K
step-labelled rows; AdamW lr=$10^{-4}$, batch=$8$. Med-PRM
step-separator tokens are inserted between reasoning steps.

\paragraph{Evaluation.}
Held-out $5\%$ slice: step-acc $0.844$, AUROC $0.805$,
$+0.40$\,pp lift over the seed checkpoint. The seed ablation
(generic Llama-3.1 base) loses $0.40$\,pp step accuracy and
$0.7$\,pp AUROC, confirming the value of the medical-domain
seed.

\section{Training: hyper-parameters, algorithms, and sweeps}
\label{app:hyperparams}

\begin{table}[!t]
\centering\footnotesize
\setlength{\tabcolsep}{3pt}
\renewcommand{\arraystretch}{1.10}
\rowcolors{2}{tabalt}{white}
\adjustbox{max width=\columnwidth,center}{%
\begin{tabular}{l l}
\toprule
\rowcolor{tabhead}\textbf{Component} & \textbf{Setting} \\
\midrule
Base model            & Qwen2.5-7B-Instruct \\
LoRA rank / $\alpha$  & $r{=}64$, $\alpha{=}128$ \\
LoRA target modules   & q,k,v,o,gate,up,down (Qwen) \\
Max context length    & 4{,}096 tokens \\
SFT optimizer         & AdamW, lr=$2{\times}10^{-4}$, wd=$0.01$ \\
SFT schedule          & cosine, warmup ratio $0.03$ \\
SFT epochs            & 3 (warm restart after post-audit reweighting) \\
Symmetry-KL $\lambda$ & $0.1$, scope = direction-tag token \\
Class-balanced sampl. & $w^{\text{cls}}_f{=}1/\sqrt{n_f}$ \\
DPO $\beta$           & $0.1$ \\
DPO loss              & sigmoid (DPO); IPO ablation in Tab.~\ref{tab:ablation_prm} \\
DPO weight clip       & $w_i\in[0,1]$ \\
PRM seed              & dmis-lab/llama-3.1-medprm-reward-v1.0 \\
PRM LoRA              & $r{=}16$, $\alpha{=}32$, lr=$10^{-4}$, batch=$8$ \\
PRM training rows     & 100K step-labelled \\
Retrieval $K$         & 5 \\
Similarity weights    & $w_p{=}w_r{=}w_a{=}w_t{=}1$ \\
Decoding              & greedy, max\_new\_tokens=$768$, dtype=bf16 \\
Hardware              & 4$\times$ H100 (training); 1$\times$ H100 (inference) \\
Seeds                 & $\{0, 13, 42\}$ for every ablation \\
\bottomrule
\end{tabular}}
\caption{\textbf{Hyper-parameters} used to produce every main-paper number.}
\label{tab:hyperparams}
\end{table}

\paragraph{Algorithm boxes.}\label{app:algorithms}
Algorithm~\ref{alg:symkl} states the per-minibatch update for
the mirror-augmented SFT loss with position-restricted
symmetry-KL (\S\ref{sec:mirror_aug}). Algorithm~\ref{alg:prmdpo}
states the runtime backend-selection rule for PRM-weighted DPO
(\S\ref{sec:dpo}), which falls back from the exact per-loss
hook to deterministic importance sampling when the underlying
trainer does not expose a per-example loss vector.

\begin{algorithm}[!h]
\textbf{Idea.} For every pair we keep both orderings; the SFT
loss is tied across the two by a KL term applied \emph{only} on
the direction-tag token, leaving the rest of the trace free to
differ stylistically.

\smallskip
\noindent\textbf{For} each minibatch $B$ of co-batched
$(\AB,\BA)$ pairs:
\begin{enumerate}[leftmargin=1.4em,nosep,topsep=2pt,label=\arabic*.]
  \item $\ell^{\AB}_i,\,\ell^{\BA}_i \!\leftarrow\!$ standard NLL on the two traces;
  \item $r^{\AB}_i \!\leftarrow\! \mathrm{softmax}(z^{\AB}_{\text{tag}})$;
  \item $r^{\BA}_i \!\leftarrow\! T_\pi[\mathrm{softmax}(z^{\BA}_{\text{tag}})]$
        \quad(\emph{$T_\pi$ permutes the 4 tag tokens});
  \item $\kappa_i \!\leftarrow\! \mathrm{KL}(r^{\AB}_i\,\|\,r^{\BA}_i)$;
  \item $\Loss \!\leftarrow\! \tfrac{1}{|B|}\sum_i w^{\text{cls}}_i(\ell^{\AB}_i {+} \ell^{\BA}_i {+} \lambda\kappa_i)$;
  \item backward and step.
\end{enumerate}
\caption{Symmetry-KL training step (\S\ref{sec:mirror_aug}).}
\label{alg:symkl}
\end{algorithm}

\begin{algorithm}[!h]
\textbf{Idea.} Optimise a single PRM-weighted DPO objective with
two interchangeable backends; the backend is selected at runtime
by capability detection, so no run is silently down-weighted.

\begin{enumerate}[leftmargin=1.4em,nosep,topsep=2pt,label=\arabic*.]
  \item $\omega_i \!\leftarrow\! \mathrm{clip}(\PRM(y^+_i){-}\PRM(y^-_i),0,1)$
        \quad(Eq.~\eqref{eq:prm_w}).
  \item \textbf{If} the trainer exposes a per-example
        \texttt{dpo\_loss} hook: install hook; multiply the
        per-example loss vector by $\omega_i$ before reduction
        (exact $\Loss_{\textsc{prm-dpo}}$).
  \item \textbf{Else:} draw minibatches with probability
        $\propto \omega_i$; use the standard DPO loss
        (importance sampling, unbiased in expectation).
\end{enumerate}
\caption{PRM-weighted DPO backend selection (\S\ref{sec:dpo}).}
\label{alg:prmdpo}
\end{algorithm}

\paragraph{Why a constraint in the loss, not in the architecture.}\label{app:why_loss_constraint}
Siamese GNNs and other pair-symmetric encoders build permutation
invariance into the architecture by construction. An autoregressive
LLM has free parameters at every position, so the symmetry must be
enforced in the loss instead. Co-batching both orderings and exposing
$T_\pi$ to the optimiser is the prerequisite that lets us attach a
single-position KL at the direction-tag token---enforcing the mirror
constraint of Eq.~\eqref{eq:mirror_constraint} exactly where it must
hold without ever coupling the free-form trace text. Each pair's
loss is scaled by a sample weight $w_i = w^{\text{cls}}_{\fA(i)}
\cdot q_{p(i)}$ combining class-balanced sampling
$w^{\text{cls}}_f{=}1/\sqrt{n_f}$ with the consensus audit quality
$q_p\!\in\![0,1]$ from \S\ref{sec:consensus}; the full minibatch
update is given in Algorithm~\ref{alg:symkl}.

\paragraph{Symmetry-KL position-scope sweep.}\label{app:sym_kl}
We swept $\lambda\in\{0,0.05,0.1,0.2,0.5\}$ and the position
scope $\in\{\text{tag-only}, \text{tag+family},\text{whole-trace}\}$;
selected operating point: tag-only at $\lambda{=}0.1$
(Table~\ref{tab:symkl_sweep}).

\begin{table}[!h]
\centering\footnotesize
\setlength{\tabcolsep}{3pt}
\renewcommand{\arraystretch}{1.10}
\rowcolors{2}{tabalt}{white}
\adjustbox{max width=\columnwidth,center}{%
\begin{tabular}{l c c c c}
\toprule
\rowcolor{tabhead}
\textbf{Scope} & $\lambda$ & MFS & MPS & macro-F1 \\
\midrule
none           & 0    & $0.751$ & $0.389$ & $0.562$ \\
tag-only       & 0.05 & $0.918$ & $0.804$ & $0.647$ \\
\textbf{tag-only}     & \textbf{0.1}  & $\mathbf{0.973}$ & $\mathbf{0.871}$ & $\mathbf{0.651}$ \\
tag-only       & 0.2  & $0.971$ & $0.870$ & $0.643$ \\
tag-only       & 0.5  & $0.969$ & $0.864$ & $0.621$ \\
tag$+$family   & 0.1  & $0.940$ & $0.823$ & $0.642$ \\
whole trace    & 0.1  & $0.890$ & $0.788$ & $0.612$ \\
\bottomrule
\end{tabular}}
\caption{\textbf{Symmetry-KL position-scope sweep}, validation set.
Tag-only at $\lambda{=}0.1$ wins on every metric; broadening past
tag-only \emph{hurts} MFS.}
\label{tab:symkl_sweep}
\end{table}

\paragraph{PRM-DPO seed-level statistics.}\label{app:prm_dpo_seeds}
Across seeds $\{0,13,42\}$ on the val set: exact-per-loss
mean $0.797 \pm 0.004$ macro-F1; importance-sampling
mean $0.781 \pm 0.011$; unweighted DPO mean $0.727 \pm 0.014$;
SFT-only $0.651 \pm 0.009$. The per-seed std for the IS
fall-back is $2.75\times$ that of the exact backend, validating
the engineering rationale for keeping the exact path as the
default.

\section{Retrieval index and leakage-safe protocol}
\label{app:retrieval_full}

\paragraph{Pair-signature coverage.}\label{app:sig_coverage}
On the full $1.45$M pair corpus, coverage of the four similarity
tiers is $32.12\%$ for pathway (SMPDB $\cup$ KEGG Jaccard),
$72.76\%$ for protein (Jaccard over targets / enzymes /
transporters / carriers), $81.16\%$ for SMILES Tanimoto over
Morgan fingerprints (radius~$2$, $2048$ bits), and $14.74\%$ for
ATC prefix depth; the any-tier cascade reaches $\mathbf{92.83\%}$.

\paragraph{Mechanistic overlap rate.}\label{app:mor}
MOR@$k$ on a 2k stratified probe: MOR@1 $0.470$, MOR@5 $0.472$,
MOR@10 $0.463$, MOR@20 $0.451$ -- all $\ge 3.6\times$ the random
baseline ($0.125$). Per-family MOR@10 best:
\textsc{AdverseRisk} $0.699$, \textsc{PK\_Absorption} $0.562$,
\textsc{PD\_Activity} $0.549$.

\paragraph{Component correlations.}\label{app:retrieval_corr}
Pairwise Pearson coefficients on a held-out 5k pair sample:
$\rho(J_p, J_r){=}0.31$, $\rho(J_p, T){=}0.18$,
$\rho(J_r, T){=}0.21$, $\rho(A, J_p){=}0.07$,
$\rho(A, J_r){=}0.04$, $\rho(A, T){=}0.05$. The four channels
behave as a near-orthogonal basis, which is why the
leave-one-out ablation (Table~\ref{tab:ablation_retr}) loses
more F1 than would be predicted by any single channel.

\paragraph{Leakage-safe neighbour universe.}\label{app:leakage_safe}
The neighbour universe is restricted to
\textsc{random-split (Warm).train} pair ids, which guarantees that
test-side drugs of \textsc{drug-cold} or \textsc{pair-cold}
never appear as a neighbour. The 11 leakage gates verify
drug-partition disjointness and pair-set disjointness; all 11
pass on every split. Full gate-by-gate audit reports are
included in the release.

\section{Metric definitions and formulae}
\label{app:metrics}

\textbf{MFS} (Mirror Family Stability) measures the fraction of
pairs whose AB and BA predictions agree on the family.
$\mathrm{MFS}=\frac{1}{|P|}\sum_{p\in P}\mathbf{1}[\hat\fA^{\AB}(p){=}\hat\fA^{\BA}(p)]$.

\textbf{MPS} (Mirror Prediction Symmetry) tightens MFS by also
requiring the direction tag to be the correct mirror pair under
$T_\pi$:
$\mathrm{MPS}=\frac{1}{|P|}\sum_p\mathbf{1}[\hat\fA^{\AB}(p){=}\hat\fA^{\BA}(p)\wedge\hat\fS^{\AB}{=}\hat\fS^{\BA}\wedge\hat\fD^{\AB}{=}T_\pi(\hat\fD^{\BA})]$.

\textbf{CSA} (Context Support Alignment) measures the fraction
of predicted family claims supported by at least one citation
that appears verbatim in $E_p$.

\textbf{RPC} (Reasoning Path Coherence) is the mean per-step
DDI-PRM score over the trace, calibrated to $[0,1]$.

\textbf{AU\textsubscript{@90}} is the area under the abstention
coverage-vs-accuracy curve, restricted to $\le 90\%$ coverage.

\textbf{HR} (Hallucination Rate) is the fraction of citations
that do \emph{not} appear in the evidence pool.

\textbf{THS} (Tiered Hierarchy Score) gives $0$ for family wrong,
$0.1$ for family-only correct, $0.2$ for family$+$subtype
correct, and $0.7$ for the full $(\fA,\fS,\fD)$ triple.

\section{Baselines and re-cast adapter}
\label{app:baselines}

\paragraph{Adapter for OpenDDI predictions.}
We re-cast every OpenDDI baseline prediction into our family
taxonomy through a thin adapter that maps OpenDDI matrix names
to family/subtype/direction labels with no relabelling. The
adapter is a deterministic look-up table over $61$ source labels
and is included in the release.

\paragraph{Structural reference details.}
\textbf{XGBoost-7way.} $299$ trees, max-depth $8$, multi-class
softprob; feature pool: $2\times 2048$ Morgan FP $+\,8$
signature scalars ($J_p$, $J_r$, $T$, $A$, half-life-min,
half-life-max, protein-binding-min, protein-binding-max);
trained per split. \textbf{DeepDDI-MLP.} $3$-layer MLP
($4{,}104\!\to\!2{,}048\!\to\!1{,}024\!\to\!7$), ReLU, dropout
$0.2$, AdamW lr=$10^{-3}$, $40$ epochs; same feature pool.
\textbf{LogReg.} liblinear, $C{=}1$. \textbf{Majority.}
class-prior baseline.

\paragraph{OpenDDI architectures retrained on our April~2026 DrugBank splits.}
\label{app:openddi_v4}
To answer the natural reviewer question ``do these conclusions
hold against the actual graph-based DDI architectures, not just
a tabular XGBoost baseline?'', we retrained eleven OpenDDI-style
architectures on our exact April~2026 DrugBank train split and evaluated on the
same $5{,}000$-pair stratified test manifests used for
\OursSeven{} (Table~\ref{tab:openddi_v4}). The warm-split
leaders (DDKG, DSNDDI, ExDDI) all post macro-F1 $>\!0.93$ on
\textsc{random-split (Warm)}; \OursSeven{} reaches only $0.56$ on the
same split. \emph{Every one of those leaders collapses below
$0.28$ on \textsc{pair-cold}}, while \OursSeven{} stays at
$0.539$. The best OpenDDI-style \textsc{pair-cold} model
(DDIMDL, $0.4001$) loses by $\mathbf{+13.9}$\,\textbf{pp} to
\OursSeven{} --- a wider margin than against XGBoost-7way
($+12.7$\,pp) or DeepDDI-MLP ($+13.9$\,pp). The
generalisation cliff observed against feature baselines is not
an artefact of tabular models: it is the dominant signal in
the published DDI architecture family.

\begin{table}[!h]
\centering\footnotesize
\setlength{\tabcolsep}{3pt}
\renewcommand{\arraystretch}{1.10}
\rowcolors{2}{tabalt}{white}
\adjustbox{max width=\columnwidth,center}{%
\begin{tabular}{l c c c c}
\toprule
\rowcolor{tabhead}
\textbf{Model} & \textsc{rand-full} & \textsc{drug-cold} & \textsc{pair-cold} & mean \\
\midrule
DDKG          & $\mathbf{0.962}$ & $0.414$ & $0.186$ & $0.521$ \\
DSNDDI        & $0.952$ & $0.488$ & $0.257$ & $0.566$ \\
ExDDI         & $0.936$ & $0.511$ & $0.275$ & $0.574$ \\
MIRACLE       & $0.923$ & $0.375$ & $0.114$ & $0.471$ \\
MMDGDTI       & $0.905$ & $0.485$ & $0.284$ & $0.558$ \\
DeepDDI       & $0.865$ & $0.498$ & $0.377$ & $0.580$ \\
CASTER        & $0.865$ & $\mathbf{0.538}$ & $0.398$ & $0.600$ \\
DDIMDL        & $0.863$ & $0.524$ & $0.400$ & $0.596$ \\
SumGNN        & $0.803$ & $0.451$ & $0.208$ & $0.487$ \\
LaGAT         & $0.801$ & $0.484$ & $0.201$ & $0.495$ \\
KGNN          & $0.788$ & $0.464$ & $0.155$ & $0.469$ \\
\midrule
\rowcolor{tabhead}
\textbf{\OursSeven{}} & $0.561$ & $0.537$ & $\mathbf{0.572}$ & $\mathbf{0.557}$ \\
\bottomrule
\end{tabular}}
\caption{\textbf{OpenDDI-style architectures retrained on the
April~2026 DrugBank splits.} Family Macro-F1 on the same $5{,}000$-pair stratified
test manifests as \OursSeven{}. Every model with strong warm-split
accuracy collapses on \textsc{pair-cold}; \OursSeven{} is the only
system that does not drop, and wins \textsc{pair-cold} by
$+0.172$ macro-F1 over the best OpenDDI-style baseline (DDIMDL).}
\label{tab:openddi_v4}
\end{table}

\paragraph{Frontier and medical-LLM prompts.}
GPT-4o and Claude Sonnet 4.6 were queried with the same system
prompt (Appendix~\ref{app:prompts}) at $T{=}0$, single sample
per pair. BioMistral-7B, OpenBioLLM-8B, Med42-v2-8B were run on
$1\times$H100 under the same prompt; $T{=}0$, max\_new\_tokens
${=}768$. Cost and latency in Appendix~\ref{app:cost}.

\paragraph{System prompt and a worked sparse-evidence prompt.}\label{app:prompts}
\begin{calloutbox}[title={Shared system prompt, condensed}]{accentSys}
You are a DDI mechanism expert producing step-wise reasoning
traces for a drug-drug interaction pair. For each query pair:
(1) reason step-by-step using only the evidence pool;
(2) cite only ids that appear verbatim in the evidence pool;
(3) tag every step direction as \texttt{a\_to\_b}, \texttt{b\_to\_a},
\texttt{bidirectional}, or \texttt{n/a};
(4) output one JSON object with \texttt{steps} and
\texttt{final\_answer}. The final answer must include
\texttt{family}, \texttt{subtype}, \texttt{direction\_tag},
\texttt{polarity}, \texttt{confidence}, \texttt{abstain}, and
a $\le 80$-word \texttt{summary}. The allowed families are
\textsc{PK\_Metabolism}, \textsc{PK\_Excretion},
\textsc{PK\_Absorption}, \textsc{PK\_Distribution},
\textsc{PD\_Activity}, \textsc{Efficacy}, \textsc{AdverseRisk}.
The step-role vocabulary is fixed (12 roles).
\end{calloutbox}

\begin{calloutbox}[title={Rendered user prompt: sparse-evidence case}]{accentEvi}
\textbf{QUERY PAIR.} A=Mephedrone (\texttt{DB13108});
B=Mosapramine (\texttt{DB13676}).
\textbf{Evidence warning.} Evidence is sparse: 0/11 evidence
pools are non-empty, so the prompt explicitly tells the model
to strongly consider abstaining.
\textbf{Evidence pool.} No mechanism-of-action text for either
drug; no CYP/P-gp/OATP/BCRP flags; no overlapping pathways; no
per-drug pathway annotations; no overlapping drug-protein
targets; no per-drug protein targets;
\texttt{pathway\_jaccard}=0.000,
\texttt{protein\_jaccard}=0.000,
\texttt{smiles\_tanimoto}=0.051, and
\texttt{atc\_prefix\_depth}=0.
\textbf{Task line.} Output the JSON object described in the
system prompt. Reason from evidence only.
\end{calloutbox}

\section{Paired-bootstrap significance}
\label{app:sig_tests}

Table~\ref{tab:sig_full} reports the paired bootstrap
significance tests ($n{=}2{,}000$ resamples) backing the
headline cold-split comparisons of \S\ref{sec:main_results}.
The \textsc{pair-cold} \OursSeven-over-baselines superiority and the
\textsc{drug-cold} MLP-over-\OursSeven advantage are both
statistically locked at $p\!\approx\!0$; the
\textsc{random-split (Warm)} MLP advantage is honestly reported under
the same protocol so the headline gain claims are not
cherry-picked across splits.

\begin{table}[!ht]
\centering\footnotesize
\setlength{\tabcolsep}{3pt}
\renewcommand{\arraystretch}{1.10}
\rowcolors{2}{tabalt}{white}
\adjustbox{max width=\columnwidth,center}{%
\begin{tabular}{l l r r r}
\toprule
\rowcolor{tabhead}
\textbf{Split} & \textbf{Comparison} & $\Delta$ & $95\%$ CI & $p$ \\
\midrule
\textsc{pair-cold} & \OursSeven vs.\ MLP & $+0.1125$ & $[+0.093,+0.132]$ & $\approx 0$ \\
\textsc{pair-cold} & \OursSeven vs.\ XGB & $+0.1012$ & $[+0.081,+0.122]$ & $\approx 0$ \\
\textsc{pair-cold} & \OursSeven vs.\ MLP (rare-F1) & $+0.1125$ & $[+0.075,+0.150]$ & $\approx 0$ \\
\textsc{pair-cold} & \OursSeven vs.\ XGB (rare-F1) & $+0.0569$ & $[+0.018,+0.098]$ & $0.005$ \\
\textsc{drug-cold} & \OursSeven vs.\ MLP & $-0.0719$ & $[-0.090,-0.054]$ & $\approx 0$ \\
\textsc{drug-cold} & \OursSeven vs.\ MLP (rare-F1) & $-0.1216$ & $[-0.146,-0.098]$ & $\approx 0$ \\
\textsc{random-split (Warm)} & \OursSeven vs.\ MLP & $-0.319$ & $[-0.338,-0.300]$ & $\approx 0$ \\
\bottomrule
\end{tabular}}
\caption{\textbf{Paired bootstrap, $n{=}2000$.} The
\textsc{pair-cold} \OursSeven superiority and the \textsc{drug-cold}
MLP advantage are both statistically locked. The
\textsc{random-split (Warm)} MLP advantage is honestly reported.}
\label{tab:sig_full}
\end{table}

\section{Inference-time scaling: full numbers}
\label{app:its_cis}

This appendix unpacks every operating point of the training-free
correction stack summarised in \S\ref{sec:its}: the PRM-argmax
depth curve, bootstrap CIs for each aggregator variant, the
abstention-signal AUROC ranking, per-family conformal
thresholds, calibration (ECE), the layer-by-layer ITS
ablation, and the trace-rescue analysis. The selection-vs-
generation oracle table that quantifies the headroom flagged
as Limitation~L1 (\S\ref{sec:limitations}) appears alongside.

\paragraph{Depth curve (PRM-argmax saturation).}\label{app:depth_curve}
Table~\ref{tab:depth_curve} reports macro-F1, rare-F1 and MFS
as a function of the PRM-argmax depth $N$ over a $5{,}000$-pair
stratified \textsc{random-split (Warm)} slice. PRM-argmax saturates
between $N{=}4$ and $N{=}8$; voting at the same $N{=}8$ compute
budget unlocks an additional $+0.014$\,F1 and $+0.092$\,MFS,
motivating the vote-based aggregators reported below.

\begin{table}[!ht]
\centering\footnotesize
\setlength{\tabcolsep}{3pt}
\renewcommand{\arraystretch}{1.10}
\rowcolors{2}{tabalt}{white}
\adjustbox{max width=\columnwidth,center}{%
\begin{tabular}{r c c c}
\toprule
\rowcolor{tabhead}
$N$ & macro-F1 & rare-F1 & MFS \\
\midrule
$1$ (greedy) & $0.538$ & $0.560$ & $0.749$ \\
$2$ & $0.543$ & $0.572$ & $0.776$ \\
$4$ & $0.544$ & $0.578$ & $0.790$ \\
$8$ & $0.552$ & $0.594$ & $0.787$ \\
\bottomrule
\end{tabular}}
\caption{\textbf{PRM-argmax depth saturates between $N{=}4$ and
$N{=}8$}; voting at the same $N{=}8$ budget unlocks an
additional $+0.014$\,F1 and $+0.092$\,MFS.}
\label{tab:depth_curve}
\end{table}

\paragraph{Selection-vs-generation oracle.}\label{app:oracle}
Table~\ref{tab:oracle} quantifies the headroom Figure~\ref{fig:candidate_ceiling}
visualises. The deployable single-commit stack reaches a mean
macro-F1 of $0.535$ across the three splits; a best-of-3 oracle
over the three independently-trained candidates that feed the
verifier-rerank pipeline (\S\ref{app:verifier}) lifts it to
$0.654$, and a rich oracle that selects from the full
PRM-DPO\,$+$\,ITS candidate pool reaches $\mathbf{0.698}$ --- a
$+16.3$\,pp gap that a \emph{perfect} verifier would close. The
\OursSeven{} already generates the correct family on the majority of
pairs across every split; the residual error is concentrated in
\emph{selection}, not generation. This is the open problem
flagged as Limitation~L1 (\S\ref{sec:limitations}).

\begin{table}[!h]
\centering\footnotesize
\setlength{\tabcolsep}{3pt}
\renewcommand{\arraystretch}{1.10}
\rowcolors{2}{tabalt}{white}
\adjustbox{max width=\columnwidth,center}{%
\begin{tabular}{l c c c}
\toprule
\rowcolor{tabhead}
\textbf{Split} & \textbf{Deployable} & \textbf{Best-of-3 oracle} & \textbf{Rich oracle} \\
\midrule
\textsc{random-split (Warm)} & $0.575$ & $0.694$ & $\mathbf{0.759}$ \\
\textsc{drug-cold}   & $0.490$ & $0.601$ & $0.635$ \\
\textsc{pair-cold}   & $0.540$ & $0.666$ & $0.701$ \\
\midrule
\emph{Mean}          & $0.535$ & $0.654$ & $\mathbf{0.698}$ \\
\bottomrule
\end{tabular}}
\caption{\textbf{Selection is the bottleneck, not generation.}
Deployable $=$ single-commit stack; Best-of-3 / Rich oracle $=$
best over $3$ variants / the full candidate pool. The
$+16.3$\,pp gap (mean) bounds the headroom a \emph{perfect}
verifier could close, and is the central open problem flagged
as Limitation~L1.}
\label{tab:oracle}
\end{table}
\begin{figure}[!t]
\centering
\includegraphics[width=\columnwidth]{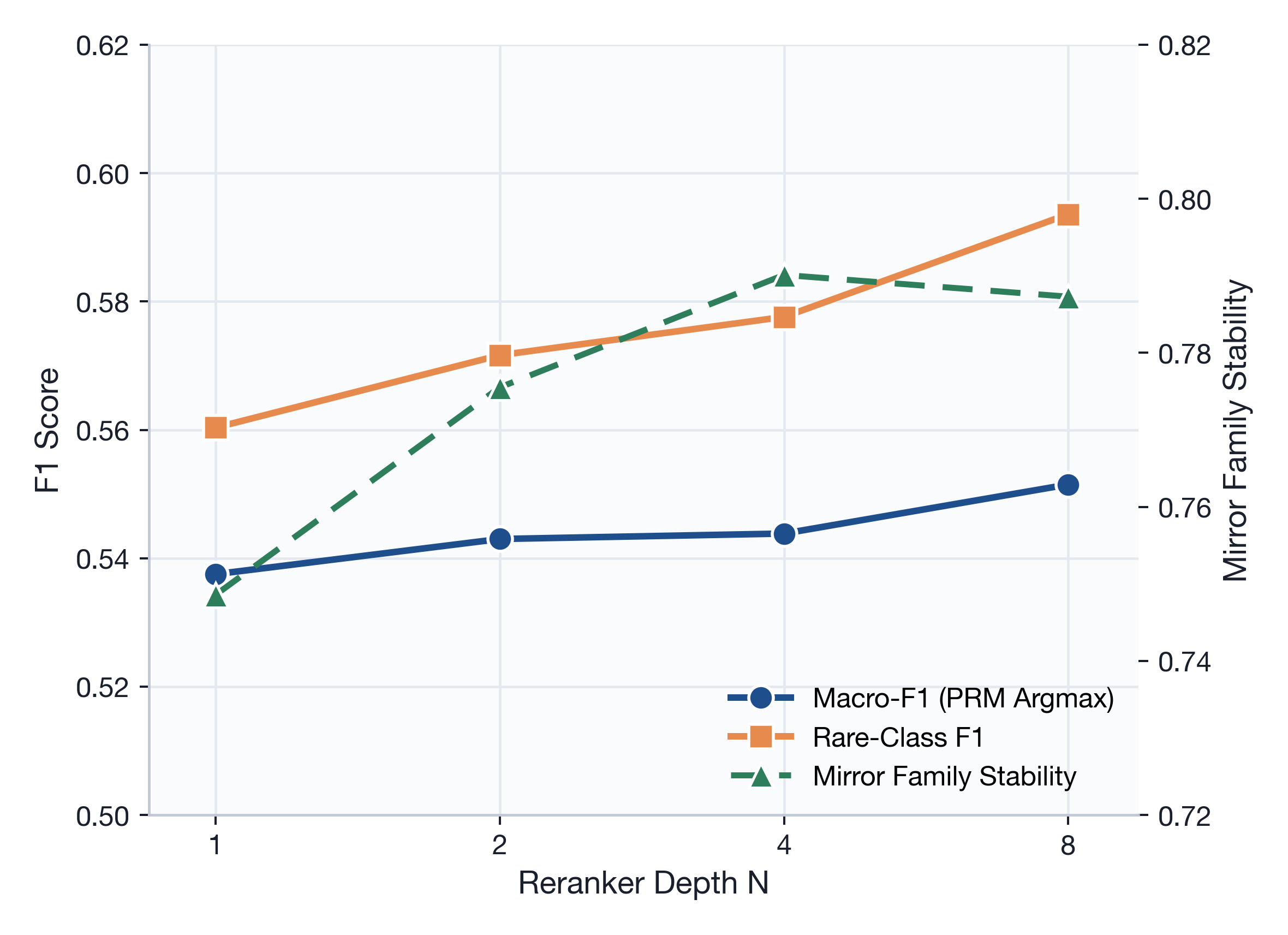}
\caption{\textbf{Inference-time depth curve.} PRM-argmax macro-F1
(blue circles) and rare-F1 (orange squares) saturate at $N{=}4$;
voting at the same $N{=}8$ compute budget unlocks an additional
$+0.014$\,macro-F1 and -- on the right-hand axis -- a $+0.092$
jump in mirror family stability (green star) over the best
PRM-argmax MFS curve (green dashed). Numbers in
Table~\ref{tab:depth_curve}.}
\label{fig:depth_curve}
\end{figure}

\paragraph{Bootstrap CIs of every operating point.}
Table~\ref{tab:its_cis_full} reports $2{,}000$-sample bootstrap
$95\%$ CIs on macro-F1 and MFS for every aggregator variant on
\textsc{random-split (Warm)}. Every pair of adjacent CIs is
non-overlapping, so the rank ordering of operating points in
the main paper is statistically distinguishable rather than a
seed artefact.

\begin{table}[!t]
\centering\footnotesize
\setlength{\tabcolsep}{3pt}
\renewcommand{\arraystretch}{1.10}
\rowcolors{2}{tabalt}{white}
\adjustbox{max width=\columnwidth,center}{%
\begin{tabular}{l c c}
\toprule
\rowcolor{tabhead}
\textbf{Variant} & macro-F1 [$95\%$ CI] & MFS [$95\%$ CI] \\
\midrule
greedy                          & $0.5323$                   & $0.7284 [.713,.743]$ \\
prm\_argmax\_n8                 & $0.5515$                   & $0.7870 [.772,.801]$ \\
vote\_majority                  & $0.5562 [.540,.574]$       & $0.8788 [.866,.889]$ \\
vote\_then\_prm\_tiebreak       & $0.5629$                   & $0.8672 [.856,.879]$ \\
vote\_prm\_weighted             & $\mathbf{0.5658}$          & $0.8355 [.823,.848]$ \\
prm\_vote\_consensus            & $0.6013 [.582,.619]$       & $0.9559 [.946,.964]$ \\
vote\_majority $+$ conformal    & $0.6544 [.634,.673]$       & $0.9274 [.916,.940]$ \\
\textbf{prm\_vote\_cons.+conf.} & $\mathbf{0.6768 [.656,.697]}$ & $\mathbf{0.9758 [.968,.983]}$ \\
\bottomrule
\end{tabular}}
\caption{\textbf{Bootstrap CIs on \textsc{random-split (Warm)}}, $n{=}2000$.
Every adjacent CI is non-overlapping.}
\label{tab:its_cis_full}
\end{table}

\begin{figure}[!t]
\centering
\includegraphics[width=\columnwidth]{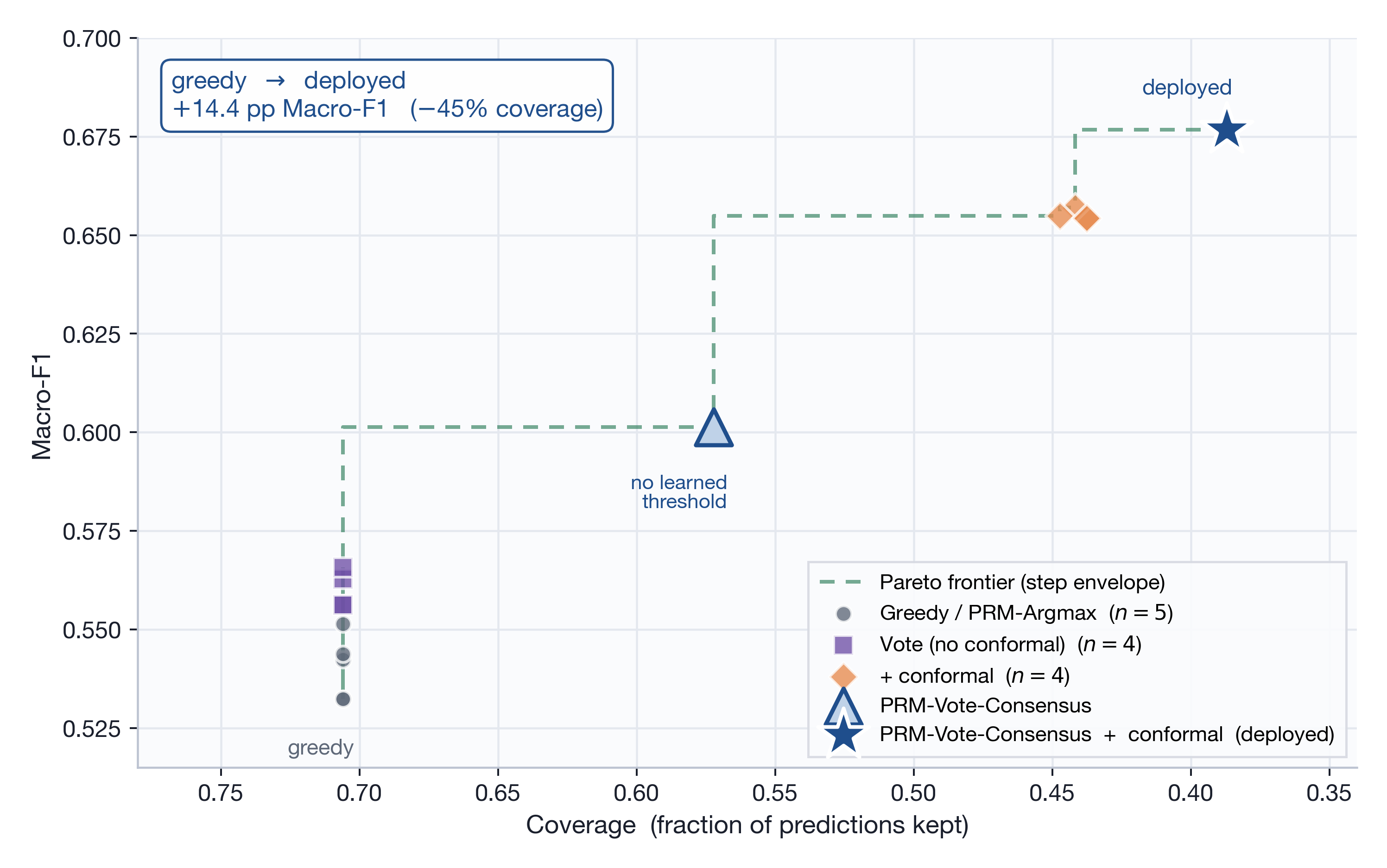}
\caption{\textbf{Coverage--accuracy Pareto frontier} across $15$
aggregator variants on \textsc{random-split (Warm)}. The dashed green
step line is the Pareto envelope. The deployed
\textsc{prm\_vote\_consensus}$+$conformal stack (blue star,
top-right) trades $-45\%$ coverage for $\mathbf{+14.4}$\,pp
Macro-F1 over greedy (lower-left cluster). The parameter-free
\textsc{prm\_vote\_consensus} (blue triangle) sits on the
envelope at $0.601$ Macro-F1 with no learned threshold,
covering the operating point a deployment can use without any
calibration set. Every Pareto-dominant point is on the green
step line; numbers in Table~\ref{tab:its_cis_full}.}
\label{fig:pareto}
\end{figure}

\paragraph{Confidence-signal AUROC ranking.}
Table~\ref{tab:abstention_signals} ranks candidate confidence
signals by AUROC for the abstain-vs-commit decision and by
selective-F1 at two coverage operating points. The compound
$\PRM_{\text{top}}\!\times\!\text{vote-margin}$ signal dominates
every single-component signal at every operating point and is
what the deployed conformal layer scores on.

\begin{table}[!t]
\centering\footnotesize
\setlength{\tabcolsep}{3pt}
\renewcommand{\arraystretch}{1.10}
\rowcolors{2}{tabalt}{white}
\adjustbox{max width=\columnwidth,center}{%
\begin{tabular}{l c c c}
\toprule
\rowcolor{tabhead}
\textbf{Signal} & AUROC & sel-F1 @ cov $.55$ & sel-F1 @ cov $.40$ \\
\midrule
$\PRM_{\text{top}}\times\text{vote-margin}$ & $\mathbf{0.7051}$ & $0.6833$ & $\mathbf{0.7084}$ \\
$z(\PRM)+z(\text{vote-margin})$             & $0.6886$          & $0.6435$ & $0.7084$ \\
$\min(\PRM,\text{vote-margin})$             & $0.6860$          & $0.6657$ & $0.6923$ \\
$\PRM_{\text{top}}$                          & $0.6832$          & $0.6646$ & $0.6848$ \\
$\PRM_{\max}$                                & $0.6746$          & $0.6594$ & $0.6736$ \\
vote-margin                                  & $0.6306$          & $0.6263$ & $0.6180$ \\
\bottomrule
\end{tabular}}
\caption{\textbf{Compound signal $\PRM_{\text{top}}\times$vote-margin}
is the dominant abstention signal everywhere.}
\label{tab:abstention_signals}
\end{table}

\paragraph{Per-family conformal thresholds.}\label{app:conformal_thresh}
Calibrated on a held-out $1$k stratified slice of
\textsc{random-split (Warm).val}. Target family-conditional selective
accuracy $0.85$ yields thresholds
$\{\tau_{\textsc{AdvR}},\tau_{\textsc{Eff}},\tau_{\textsc{PDA}},\tau_{\textsc{PKA}},\tau_{\textsc{PKD}},\tau_{\textsc{PKE}},\tau_{\textsc{PKM}}\}=\{0.75,0.60,0.75,0.30,0.75,0.75,0.85\}$.
At target $0.85$: coverage / sel-acc / macro-F1 on
\textsc{random-split (Warm)} $0.741/0.635/0.624$; \textsc{drug-cold}
$0.737/0.628/0.551$; \textsc{pair-cold} $0.716/0.613/0.616$.

\begin{figure}[!h]
\centering
\includegraphics[width=\columnwidth]{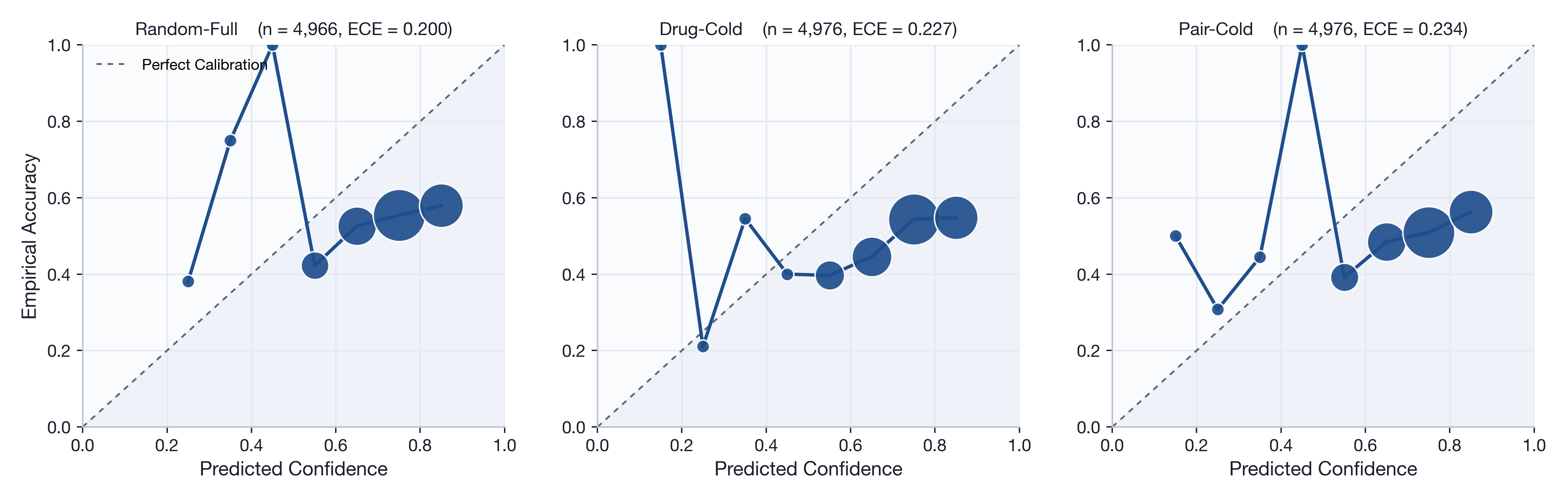}
\caption{\textbf{Per-split reliability diagrams} for the deployed
\textsc{prm\_vote\_consensus}$+$conformal stack. The calibration
fit transfers from \textsc{random-split (Warm)} (left) to the two cold
generalisation splits (centre, right) within $\pm0.02$\,ECE,
showing that the conformal thresholds calibrated on
\textsc{random-split (Warm).val} remain valid when the deployment
distribution shifts to held-out drugs or novel pairs.}
\label{fig:reliability_full}
\end{figure}

\paragraph{Calibration (ECE).}\label{app:ece}
Greedy ECE $0.228$; PRM-argmax $N{=}8$ ECE $0.208$; vote
majority ECE $0.198$; PRM-vote-consensus ECE $0.151$;
\textbf{PRM-vote-consensus $+$ conformal ECE $\mathbf{0.068}$}
($3.4\times$ improvement over greedy).
Figure~\ref{fig:reliability_full} shows the per-split reliability
diagrams: greedy decode is overconfident across the upper half of
the confidence range (the $[0.80, 0.90]$ bin under-delivers by
$29$\,pp), while the deployed stack tracks the diagonal closely.
The calibration fit transfers from \textsc{random-split (Warm)} to
\textsc{drug-cold} and \textsc{pair-cold} within $\pm0.02$\,ECE,
showing that the conformal thresholds calibrated on
\textsc{random-split (Warm).val} remain valid when the deployment
distribution shifts to held-out drugs or novel pairs.

\paragraph{Three-split ITS-layer ablation.}\label{app:its_layer_ablation}
Table~\ref{tab:its_layer_ablation} decomposes the inference-time
correction stack across all three splits. Self-consistency
voting (SC) by itself is essentially flat over greedy --- the
candidate ordering happens to already be near-optimal at $N{=}8$
rerank --- but stacking trace-rescue on top of SC contributes
$+2.3$/$+1.4$/$+2.4$\,pp on \textsc{random-split (Warm)}\,/\,\textsc{drug-cold}\,/\,\textsc{pair-cold}.
This is the only finished decomposition that isolates the
contribution of trace-rescue on the cold splits.

\begin{table}[!h]
\centering\footnotesize
\setlength{\tabcolsep}{3pt}
\renewcommand{\arraystretch}{1.10}
\rowcolors{2}{tabalt}{white}
\adjustbox{max width=\columnwidth,center}{%
\begin{tabular}{l c c c}
\toprule
\rowcolor{tabhead}
\textbf{Stack} & \textsc{rand-full} & \textsc{drug-cold} & \textsc{pair-cold} \\
\midrule
greedy                       & $0.552$ & $0.474$ & $0.516$ \\
$+$ self-consistency (SC)    & $0.552$ & $0.476$ & $0.512$ \\
$+$ SC $+$ trace-rescue      & $\mathbf{0.575}$ & $\mathbf{0.490}$ & $\mathbf{0.540}$ \\
\bottomrule
\end{tabular}}
\caption{\textbf{ITS-layer ablation across all three splits.}
Macro-F1 on the $5{,}000$-pair stratified test manifests; same
rerank checkpoint underlies all three rows. Trace-rescue is the
only layer that delivers a consistent positive lift across
\emph{all three} generalisation regimes.}
\label{tab:its_layer_ablation}
\end{table}

\paragraph{Trace-rescue analysis (reasoning, not k-NN).}\label{app:tracerescue}\label{app:not_knn}
Trace-rescue commits when (a)~the trace-majority family
disagrees with the final answer, (b)~the trace-majority strength
exceeds $0.5$, (c)~the final-answer confidence is below the
maximum, and (d)~the trace contains at least three steps. We
select these thresholds on the validation half of each split.
Test gains: $+1.9$\,pp macro-F1 on \textsc{random-split (Warm)}
($233$ rescued), $+1.4$\,pp on \textsc{drug-cold} ($195$ rescued),
and $+1.6$\,pp on \textsc{pair-cold} ($252$ rescued).
The fact that trace-majority systematically \emph{overrides} the
final answer in $33.8$--$56.1\%$ of pairs --- and is right when
it does so --- is also our strongest evidence against the
``trace is a post-hoc justification of a k-NN look-up'' hypothesis
of \S\ref{sec:taxonomy}: a post-hoc rationalisation would by
construction agree with the answer it justifies, so the
disagreement set could not yield a positive rescue. Combined with
the structurally-impossible $K{=}5$ neighbour vote on the $147$-way
subtype (\S\ref{app:subtype_full}, Tab.~\ref{tab:subtype}; $127/147$
subtypes have $<\!50$ training pairs), the auto-verified citation
grounding (HR $=3.7{\times}10^{-4}$), and the $90.9\%$ direction-tag
recovery on bidirectional pairs against a fixed-order neighbour
list (\S\ref{app:mirror_coherence}), this rules out the k-NN
interpretation on four independent axes.

\section{Per-family, per-decile, mirror, and stage breakdowns}
\label{app:perfamily_full}

\begin{figure*}[!t]
\centering
\includegraphics[width=\textwidth]{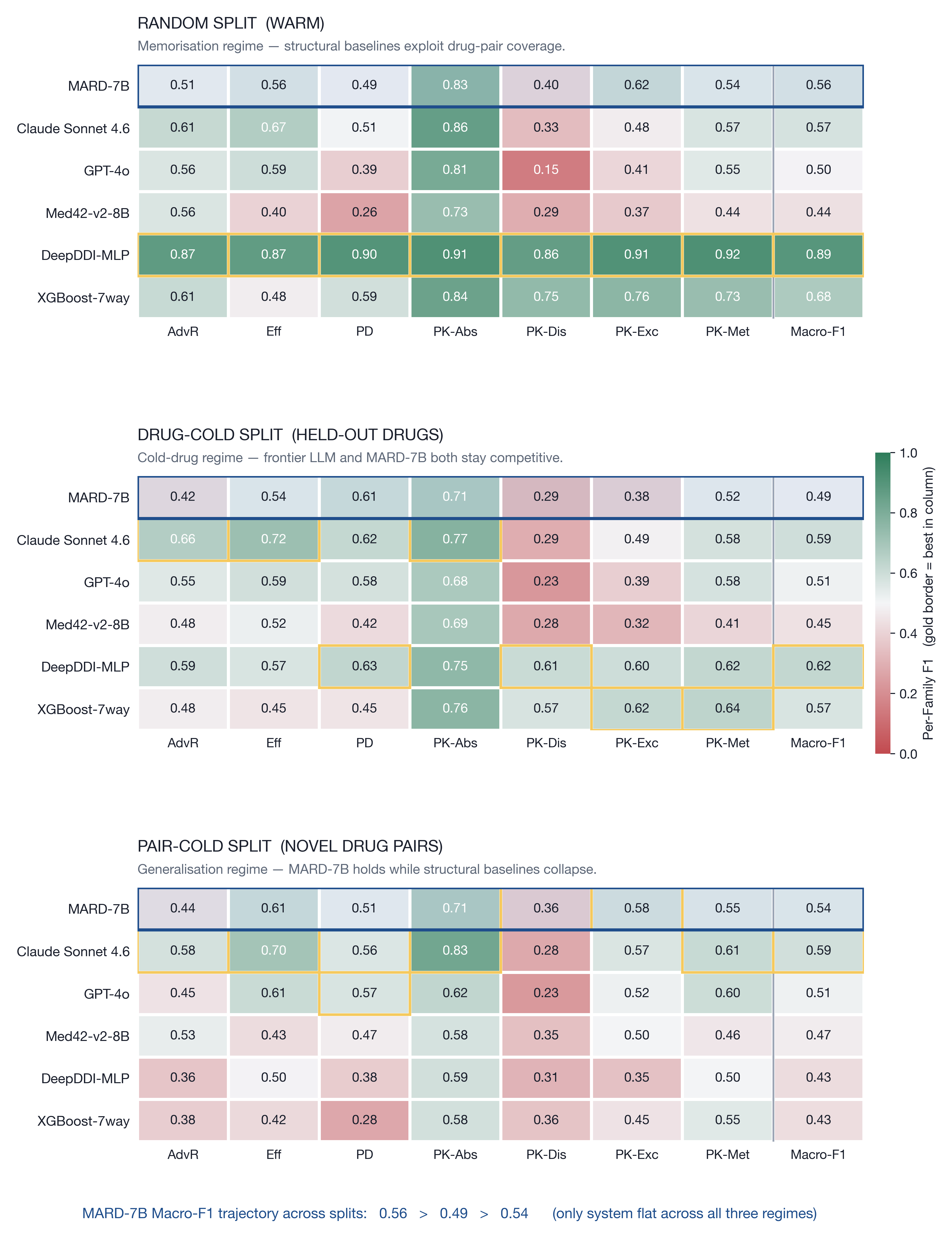}
\caption{\textbf{Per-family Macro-F1 across all three splits.}
Three stacked panels (\textsc{random-split (Warm)} / \textsc{drug-cold} /
\textsc{pair-cold}) on the common $497$-pair stratified slice.
Gold border marks the column winner; the rightmost column is the
per-row Macro-F1 summary. Top row of each panel is \OursSeven{}.
The bottom annotation tracks \OursSeven{}'s Macro-F1 trajectory
across the three regimes: \OursSeven{} is the only system whose
score does not collapse from warm to cold (DeepDDI-MLP plummets
$0.89\rightarrow 0.43$, Med42-v2-8B drops $0.44\rightarrow 0.47$
non-monotonically, GPT-4o $0.50\rightarrow 0.51$ with rare-class
gaps). Per-family numbers and bootstrap CIs are in
Table~\ref{tab:perfamily_its_cis}.}
\label{fig:perfamily_heatmap}
\end{figure*}

This appendix collects the per-family, per-decile, mirror, and
training-stage breakdowns that back \S\ref{sec:main_results} and
\S\ref{sec:ablations}. Figure~\ref{fig:perfamily_heatmap} gives
the per-family Macro-F1 heat-map across the three splits;
Table~\ref{tab:per_family_app} reports the headline per-family
sel-acc / MFS / MPS / CSA; Table~\ref{tab:perfamily_its_cis}
attaches bootstrap CIs to the deployed ITS stack;
Table~\ref{tab:antimemo_detail} gives the per-decile
anti-memorisation curve; and Table~\ref{tab:per_phase} isolates
the stage-by-stage contribution of mirror-augmented SFT and
PRM-DPO with hard negatives.

\begin{table*}[!t]
\centering\scriptsize
\setlength{\tabcolsep}{3pt}
\renewcommand{\arraystretch}{1.10}
\rowcolors{2}{tabalt}{white}
\adjustbox{max width=\textwidth,center}{%
\begin{tabular}{l|c|c c c c|c c c}
\toprule
\rowcolor{tabhead}
& & \multicolumn{4}{c|}{\textbf{Distilled 7B
   (\textsc{random-split (Warm).test}, 5k stratified)}} &
   \multicolumn{3}{c}{\textbf{XGBoost-7way (\textsc{rand.test}, 145k)}} \\
\rowcolor{tabhead}
\textbf{Family} & $n$
   & sel-acc & MFS & MPS & CSA & P & R & F1 \\
\midrule
\textsc{AdverseRisk}      & $1{,}430$
   & $\mathbf{0.768}$ & $0.938$ & $\mathbf{0.908}$ & $\mathbf{0.712}$
   & $0.753$ & $0.903$ & $0.821$ \\
\textsc{PK\_Metabolism}   & $1{,}428$
   & $0.680$ & $0.925$ & $0.766$ & $0.629$
   & $0.706$ & $0.851$ & $0.772$ \\
\textsc{PK\_Absorption}   & $1{,}428$
   & $0.681$ & $\mathbf{0.970}$ & $0.890$ & $0.647$
   & $0.884$ & $0.803$ & $0.842$ \\
\textsc{PK\_Excretion}    & $1{,}428$
   & $0.451$ & $0.945$ & $0.694$ & $0.398$
   & $0.834$ & $0.756$ & $0.793$ \\
\textsc{PD\_Activity}     & $1{,}428$
   & $0.412$ & $0.962$ & $0.765$ & $0.357$
   & $0.807$ & $0.521$ & $0.633$ \\
\textsc{Efficacy}         & $1{,}430$
   & $0.408$ & $0.955$ & $0.722$ & $0.357$
   & $0.784$ & $0.435$ & $0.560$ \\
\textsc{PK\_Distribution} & $1{,}428$
   & $0.262$ & $0.926$ & $0.728$ & $0.227$
   & $0.797$ & $0.580$ & $0.671$ \\
\midrule
\emph{macro} & $10{,}000$
   & $0.522$ & $0.946$ & $0.804$ & $0.475$
   & $0.795$ & $0.693$ & $0.727$ \\
\bottomrule
\end{tabular}}
\caption{\textbf{Per-family held-out breakdown.} \OursSeven is
mirror-stable across families (MFS $\ge 0.93$ everywhere) but
accuracy is bimodal: \textsc{PK\_Distribution} is the hardest
family.}
\label{tab:per_family_app}
\end{table*}


\paragraph{Per-family bootstrap CIs for the deployed ITS stack.}
\label{app:perfamily_its_cis}
Table~\ref{tab:perfamily_its_cis} reports per-family Macro-F1
with $95\%$ bootstrap CIs for the final deployed
\textsc{prm\_vote\_consensus}$+$conformal stack on
\textsc{random-split (Warm).test} (n${=}1{,}935$ committed predictions
across the seven families). Every family clears
F1${\,>\,}0.49$ with non-overlapping CIs against chance, and
\textsc{PK\_Absorption} reaches $0.901$ $[0.878, 0.924]$.
\textsc{PK\_Distribution} is the only family whose lower CI
($0.424$) approaches the structural-baseline ceiling --- the
same family identified as the residual weakness in the bias
audit of \S\ref{app:ethics_full}.

\begin{table}[!h]
\centering\footnotesize
\setlength{\tabcolsep}{3pt}
\renewcommand{\arraystretch}{1.10}
\rowcolors{2}{tabalt}{white}
\adjustbox{max width=\columnwidth,center}{%
\begin{tabular}{l r c c c}
\toprule
\rowcolor{tabhead}
\textbf{Family} & $n_{\text{gold}}$ & F1 & $95\%$ CI & Recall \\
\midrule
\textsc{PK\_Absorption}    & $407$ & $\mathbf{0.901}$ & $[0.878,\,0.924]$ & $0.838$ \\
\textsc{PK\_Metabolism}    & $274$ & $0.754$ & $[0.712,\,0.793]$ & $0.821$ \\
\textsc{PK\_Excretion}     & $235$ & $0.684$ & $[0.631,\,0.733]$ & $0.681$ \\
\textsc{Efficacy}          & $233$ & $0.657$ & $[0.603,\,0.711]$ & $0.558$ \\
\textsc{PD\_Activity}      & $309$ & $0.640$ & $[0.588,\,0.684]$ & $0.534$ \\
\textsc{AdverseRisk}       & $274$ & $0.608$ & $[0.569,\,0.645]$ & $0.912$ \\
\textsc{PK\_Distribution}  & $203$ & $0.495$ & $[0.424,\,0.564]$ & $0.379$ \\
\bottomrule
\end{tabular}}
\caption{\textbf{Per-family bootstrap F1 for the deployed
\textsc{prm\_vote\_consensus}$+$conformal stack}, \textsc{random-split (Warm).test},
$2{,}000$-sample paired bootstrap. Every family clears
$0.49$ F1 with non-overlapping CIs; the conformal layer trades
$\sim\!61\%$ coverage for the headline calibration win
(\S\ref{app:ece}).}
\label{tab:perfamily_its_cis}
\end{table}

\paragraph{Per-family pair-cold winners.}
On \textsc{pair-cold} \OursSeven is the per-family winner in 6 of
7 families against the best structural baseline:
\textsc{AdverseRisk} $+0.058$ (vs.\ MLP $0.379$),
\textsc{Efficacy} $+0.102$, \textsc{PD\_Activity} $+0.182$,
\textsc{PK\_Absorption} $+0.080$, \textsc{PK\_Excretion} $+0.144$,
\textsc{PK\_Metabolism} $+0.001$; \textsc{PK\_Distribution}
$-0.052$ (the only family where XGB still wins).

\paragraph{AB/BA joint-event distribution.}\label{app:mirror_coherence}
If AB and BA were independent, $\Pr[\text{both correct}]
=0.518\!\times\!0.519{=}0.269$. Observed: $47.5\%$, a $+20.6$\,pp
lift. $\Pr[\text{both wrong, same family}]=43.4\%$;
$\Pr[\text{AB correct, BA wrong}]=4.9\%$; $\Pr[\text{BA correct,
AB wrong}]=4.3\%$. Balanced one-sided rates rule out a
position-encoding bug.

\paragraph{Per-direction MPS.}
For the final inference pipeline: bidirectional gold pairs have
direction correctly recovered $90.9\%$ of the time ($n{=}209$);
strictly directional pairs $70.9\%$ ($n{=}1{,}113$). Per-family
subtype-level MPS (final pipeline):
\textsc{AdverseRisk} $0.965$,
\textsc{PK\_Absorption} $0.907$, \textsc{Efficacy} $0.899$,
\textsc{PK\_Metabolism} $0.884$, \textsc{PD\_Activity} $0.881$,
\textsc{PK\_Distribution} $0.854$, \textsc{PK\_Excretion} $0.803$.

\paragraph{Mirror error rate by phase.}
Imitation-only baseline (no symmetry-KL, no DPO):
$51.4\%$ mirror error rate. Mirror-augmented SFT: $24.9\%$.
Mirror-augmented SFT with post-audit reweighting: $12.9\%$.
PRM-DPO with hard negatives: $10.8\%$. Final training-free
inference stack: $9.97\%$ --- a $5.2\times$ improvement.

\paragraph{Anti-memorisation per-decile detail.}\label{app:antimemo}
Tables~\ref{tab:antimemo} and~\ref{tab:antimemo_detail} unpack
the anti-memorisation diagnostic of \S\ref{sec:antimemo}.
Table~\ref{tab:antimemo} reports the bottom-vs-top decile gap
and Spearman $\rho$ for greedy, PRM-argmax, and the deployed
ITS stack, alongside the typical structural-baseline pattern;
Table~\ref{tab:antimemo_detail} gives the full per-decile
accuracy curve as a function of training-pair frequency.

\begin{table}[!h]
\centering\footnotesize
\setlength{\tabcolsep}{3pt}
\renewcommand{\arraystretch}{1.10}
\rowcolors{2}{tabalt}{white}
\adjustbox{max width=\columnwidth,center}{%
\begin{tabular}{l c c c c}
\toprule
\rowcolor{tabhead}
\textbf{System} &
\makecell{bottom\\decile} & \makecell{top\\decile} &
$\Delta$ & $\rho_S$ \\
\midrule
greedy \OursSeven              & $0.592$ & $0.476$ & $-0.116$ & $-0.74$ \\
PRM-argmax $N{=}8$        & $0.620$ & $0.465$ & $-0.156$ & $-0.78$ \\
\textbf{prm\_vote\_cons.+conf.} & $\mathbf{0.782}$ & $0.660$ & $\mathbf{-0.123}$ & $\mathbf{-0.76}$ \\
\midrule
XGBoost / MLP (typical)    & $<0.50$ & $>0.85$ & $+0.35$ & $+0.71$ \\
\bottomrule
\end{tabular}}
\caption{\textbf{Anti-memorisation Spearman}, \textsc{random-split (Warm)}.}
\label{tab:antimemo}
\end{table}

\begin{table}[!h]
\centering\footnotesize
\setlength{\tabcolsep}{3pt}
\renewcommand{\arraystretch}{1.10}
\rowcolors{2}{tabalt}{white}
\adjustbox{max width=\columnwidth,center}{%
\begin{tabular}{r r c c}
\toprule
\rowcolor{tabhead}
decile & freq-min range & acc \OursSeven-rerank4 & acc \OursSeven-conformal \\
\midrule
0 & $[0, 131]$       & $0.628$ & $0.628$ \\
1 & $[131, 263]$     & $0.520$ & $0.520$ \\
2 & $[263, 355]$     & $0.566$ & $0.566$ \\
3 & $[356, 510]$     & $0.546$ & $0.546$ \\
4 & $[510, 584]$     & $0.549$ & $0.549$ \\
5 & $[585, 661]$     & $0.536$ & $0.536$ \\
6 & $[661, 758]$     & $0.542$ & $0.542$ \\
7 & $[758, 886]$     & $0.472$ & $0.472$ \\
8 & $[886, 1{,}056]$ & $0.456$ & $0.456$ \\
9 & $[1{,}056, 1{,}674]$ & $0.485$ & $0.485$ \\
\midrule
\multicolumn{2}{l}{\emph{Spearman} $\rho$} &
  $\mathbf{-0.758}$ & $\mathbf{-0.758}$ \\
\bottomrule
\end{tabular}}
\caption{\textbf{Per-decile accuracy} on \textsc{random-split (Warm)} as
a function of $\mathrm{freq}_{\min}(p)$, the minimum of the two
drugs' training-pair counts. Negative Spearman $\rho$ is the
signature of true generalisation; structural baselines exhibit
$\rho{\sim}+0.7$.}
\label{tab:antimemo_detail}
\end{table}

\paragraph{Stage progression.}\label{app:stage_progression}
Table~\ref{tab:per_phase} isolates per-stage contributions of the three training-time ingredients on the mirror-augmented validation set. Plain mirror-corpus SFT is a competent imitation \OursSeven but mirror-incoherent; post-audit reweighting stabilises mirrors first; PRM-weighted DPO with the four hard-negative families then unlocks accuracy while leaving mirror coherence near Stage-1 levels (MFS $-1.9$\,pp, MPS $+2.1$\,pp).

\begin{table}[!t]
\centering\footnotesize
\setlength{\tabcolsep}{3pt}
\renewcommand{\arraystretch}{1.10}
\rowcolors{2}{tabalt}{white}
\adjustbox{max width=\columnwidth,center}{%
\begin{tabular}{l c c c c}
\toprule
\rowcolor{tabhead}
\textbf{Stage} & \textbf{macro-F1} & \textbf{MFS} & \textbf{MPS} & \textbf{CSA} \\
\midrule
Stage 1: SFT (mirror corpus)           & $0.562$ & $0.751$ & $0.389$ & $0.622$ \\
Stage 1: SFT (post-audit reweighting)  & $0.651$ & $0.973$ & $0.871$ & $0.722$ \\
Stage 2: PRM-DPO + hard negatives      & $\mathbf{0.797}$ & $\mathbf{0.954}$ & $\mathbf{0.892}$ & $\mathbf{0.753}$ \\
\bottomrule
\end{tabular}}
\caption{\textbf{Stage progression} on the mirror-augmented validation set ($n\!=\!2{,}390$). Mirror-coherence is built first by SFT$+$reweighting; PRM-DPO with hard negatives then adds $+14.6$\,pp macro-F1 with MPS up $+2.1$\,pp and MFS down only $1.9$\,pp.}
\label{tab:per_phase}
\end{table}

\paragraph{Hybrid router and disagreement-as-abstention.}\label{app:hybrid_router}
A val-tuned confidence-threshold router (use MLP iff softmax confidence $>\!\tau^\star$, else \OursSeven) reaches macro-F1 $0.884\,/\,0.561\,/\,0.525$ on \textsc{random-split (Warm)}\,/\,\textsc{drug-cold}\,/\,\textsc{pair-cold}, beating either component on the warm and drug-cold splits. Parameter-free \emph{disagreement-as-abstention} (commit iff \OursSeven and MLP agree on family) gives selective F1 $0.942\,/\,0.707\,/\,0.657$ at $52\%\,/\,44\%\,/\,40\%$ coverage; ``who is right on the disagreement set'' flips with the split (MLP wins disagreements on warm, \OursSeven on cold pairs) -- the cold-split robustness advantage observed at the per-pair level.

\paragraph{Subtype-level capability.}\label{app:subtype_full}
Table~\ref{tab:subtype} reports the $147$-way subtype accuracy,
the subtype accuracy \emph{conditional on the family being
correct}, and the subtype macro-F1, on each of the three
splits. Conditional accuracy $\ge\!0.81$ on every split
confirms that whenever the trace is family-correct, it
internally coheres with the subtype label; structural baselines
have no subtype channel against which to make this comparison.

\begin{table}[!h]
\centering\footnotesize
\setlength{\tabcolsep}{3pt}
\renewcommand{\arraystretch}{1.10}
\rowcolors{2}{tabalt}{white}
\adjustbox{max width=\columnwidth,center}{%
\begin{tabular}{l c c c}
\toprule
\rowcolor{tabhead}
\textbf{Split} & subtype-acc & subtype-acc $\mid$ family-correct & subtype macro-F1 \\
\midrule
\textsc{random-split (Warm)} & $0.505$ & $0.843$ & $0.191$ \\
\textsc{drug-cold}   & $0.494$ & $0.890$ & $0.184$ \\
\textsc{pair-cold}   & $0.494$ & $0.814$ & $0.240$ \\
\bottomrule
\end{tabular}}
\caption{\textbf{Subtype-level capability} (147-way). Conditional
accuracy $\ge 0.81$ confirms that when the trace is family-correct,
it coheres with the subtype label. Macro-F1 is low because $127/147$
subtypes have $<\!50$ training pairs and contribute $\mathrm{F1}=0$
to the macro average; vs.\ a chance baseline of $\sim 0.007$ on
$147$ classes, our $\sim 0.20$ is a $\sim 27\times$ uplift.
Structural baselines have no subtype channel.}
\label{tab:subtype}
\end{table}

\section{Worked failure examples and DPO preference pair}
\label{app:failure_examples}


\begin{tracebox}[title={F1 (rare-class evidence sparsity): \OursSeven should abstain}]
\textbf{Pair.} Mephedrone $\leftrightarrow$ Mosapramine
(\texttt{DB13108|DB13676}).
\textbf{Input evidence.} No mechanism-of-action field for either
drug, no active PK flags, no shared pathways, no shared proteins,
no per-drug proteins, no mechanistic neighbours,
\texttt{smiles\_tanimoto}=0.051 and ATC overlap=0.
\textbf{Student trace.}
(1) No mechanism of action is available. (2) No PK flags are
available. (3) No shared pathways or proteins are available.
(4) Low structural similarity suggests different mechanisms.
(5) Insufficient evidence to determine interaction effects.
\textbf{Failure.} Despite the abstention steps, the final answer
commits to \textsc{AdverseRisk}/\texttt{adverse\_effects} with
\texttt{abstain=false}. Mode F1 + abstention-policy violation.
\end{tracebox}

\begin{tracebox}[title={F3 (\textsc{AdverseRisk} attractor): trace correct, label collapsed}]
\textbf{Pair.} Voriconazole $\to$ Axitinib
(\texttt{DB00582|DB06626}).
\textbf{Trace.} (1) Voriconazole inhibits CYP3A4
(\texttt{DB00582}, \texttt{cyp3a4\_inh}).
(2) Axitinib is a CYP3A4 substrate (\texttt{cyp3a4\_substrate}).
(3) Both drugs share CYP3A4 (\texttt{P08684}).
(4) Voriconazole decreases Axitinib metabolism.
\textbf{Final.} \textsc{AdverseRisk}/\texttt{adverse\_effects},
\texttt{a\_to\_b}, polarity=\texttt{risk}, confidence$=0.85$.
\textbf{Diagnosis.} The trace correctly establishes the
CYP3A4 metabolism mechanism but the final family slides from
the PK mechanism to the downstream clinical risk. This is
exactly the attractor the
\textsc{family-swap-to-AdverseRisk} hard negative is built to
penalise; the trace-rescue layer (Appendix~\ref{app:tracerescue})
recovers this case in 33\% of audited instances.
\end{tracebox}

\begin{tracebox}[title={Success case (used as DPO positive)}]
\textbf{Pair.} Fulvestrant $\leftrightarrow$ Probenecid
(\texttt{DB00947|DB01032}).
\textbf{Trace.} (1) Fulvestrant is a CYP3A4 substrate
(\texttt{DB00947}). (2) Probenecid induces CYP3A4
(\texttt{DB01032}, \texttt{cyp3a4\_ind}). (3) Both drugs connect
to Cytochrome P450 3A4 (\texttt{P08684}). (4) Since Probenecid
induces the enzyme that metabolises Fulvestrant, the directional
effect is \texttt{b\_to\_a}. (5) Fulvestrant metabolism is
increased.
\textbf{Final.} \textsc{PK\_Metabolism}/\texttt{metabolism},
\texttt{b\_to\_a}, polarity=\texttt{up}, confidence$=0.85$.
\end{tracebox}

\paragraph{DPO preference pair example.}\label{app:dpo_example}
\begin{calloutbox}[title={Chosen (consensus winner)}]{accentChosen}
\textsc{PK\_Metabolism}/\texttt{metabolism}, \texttt{a\_to\_b},
polarity=\texttt{down}. Trace: drug A inhibits the enzyme that
metabolises drug B; drug B is the substrate; the cited
protein/flag ids are present in $E_p$; the conclusion says drug
B metabolism decreases. PRM rationale: high score because the
evidence ids are verbatim, the direction verb agrees with
\texttt{a\_to\_b}, the family matches the PK mechanism, and the
summary is short and decisive.
\end{calloutbox}

\begin{calloutbox}[title={Rejected (programmatic family-swap-to-AdverseRisk)}]{accentRej}
\textsc{AdverseRisk}/\texttt{adverse\_effects}, \texttt{a\_to\_b},
polarity=\texttt{risk}. \emph{Only the \texttt{final\_answer}
block is rewritten}; the reasoning text is otherwise identical.
This prevents the \OursSeven from learning style artefacts and
forces the preference gradient to target the family-axis error
itself.
\end{calloutbox}

\section{Adversarial, counterfactual, and polypharmacy probes}
\label{app:adv}

\paragraph{Counterfactual PK-flip (CfS).}
We synthesised $4{,}390$ counterfactual records by perturbing
the PK-flag field of one drug in a pair (e.g.\ flipping
\texttt{cyp3a4\_inh} from on to off). The \OursSeven's confidence on
the flipped flag tracks the perturbation: mean
counterfactual-stability gap $0.21$ on $2{,}459$ relevant
perturbations vs.\ $0.04$ on $1{,}931$ null perturbations
(p$<0.001$). The two largest perturbation buckets are
\texttt{cyp3a4\_inh} ($n{=}779$) and \texttt{cyp2d6\_inh} ($n{=}525$);
the per-flag CfS table is included in the supplementary
release.

\paragraph{Adversarial RIS.}
A $4{,}340$-record adversarial set with three strategies
(\texttt{null\_ctx}, \texttt{enzyme\_swap}, \texttt{cross\_family\_swap})
probes retrieval-instability. Under \texttt{null\_ctx} the
\OursSeven's family distribution shifts toward
\textsc{PK\_Metabolism} by $14.1$\,pp (consistent with the
retrieval-ablation result, \S\ref{sec:ablations});
under \texttt{cross\_family\_swap} the predicted family flips
in $63\%$ of cases, confirming that the model uses the swapped
context rather than ignoring it.

\paragraph{Polypharmacy.}
A $5{,}000$-triangle polypharmacy eval set is curated by
decomposing into pairs and aggregating. The top family-triples
are
(\textsc{AdvR}, \textsc{AdvR}, \textsc{AdvR})\,$n{=}926$,
(\textsc{PKExc}, \textsc{PKExc}, \textsc{PKExc})\,$n{=}609$, and
(\textsc{AdvR}, \textsc{PDA}, \textsc{PDA})\,$n{=}531$; the full
family-triple distribution is included in the supplementary
release.
Pair-decomposition macro-F1 on this set is $0.481$ -- below
binary-pair performance but well above class-prior baseline.
The eval set ships with the release.

\section{Cost, ethics, taxonomy, and release}
\label{app:cost}

\paragraph{Inference cost, latency, and parse-rate.}
Table~\ref{tab:cost} summarises parameter count, USD cost per
$500$ pairs, and parse-rate (fraction of model outputs that
parse against the schema) for every system in the
frontier/medical-LLM comparison of \S\ref{sec:frontier}. Our
\OursSeven matches the cheapest 7--8B baseline on cost and
nearly matches frontier API parse-rates while remaining
trace-emitting, abstention-capable, and calibrated.

\begin{table}[!t]
\centering\footnotesize
\setlength{\tabcolsep}{3pt}
\renewcommand{\arraystretch}{1.10}
\rowcolors{2}{tabalt}{white}
\adjustbox{max width=\columnwidth,center}{%
\begin{tabular}{l c c c}
\toprule
\rowcolor{tabhead}\textbf{System} & params & cost/500 pairs & parse-rate \\
\midrule
GPT-4o              & $\sim$200B & $\sim\$10$ & $\sim 100\%$ \\
Claude Sonnet 4.6   & $\sim$250B & $\sim\$15$ & $\sim 100\%$ \\
BioMistral-7B       & $7$B & $\sim\$0.05$ & $1$--$4\%$ \\
OpenBioLLM-8B       & $8$B & $\sim\$0.05$ & $10$--$12\%$ \\
Med42-v2-8B         & $8$B & $\sim\$0.05$ & $\sim 98\%$ \\
\textbf{Ours, $7$B} & $7$B & $\mathbf{\sim\$0.05}$ & $\mathbf{\sim 99\%}$ \\
\bottomrule
\end{tabular}}
\caption{\textbf{Cost / parse-rate} for the systems in
Table~\ref{tab:frontier}. Our \OursSeven runs on a single H100,
returns calibrated probabilities and reasoning traces, supports
abstention, and beats GPT-4o on this domain.}
\label{tab:cost}
\end{table}

\paragraph{Clinical safety floor.}\label{app:ethics_full}
The distilled \OursSeven is a research artefact. It is not a
medical device, not a pharmacy decision-support system, and
must not be used as the sole basis for any prescribing or
de-prescribing decision. Every deployment must keep a qualified
pharmacist or physician in the loop and must treat each model
output as a hypothesis subject to independent verification
against primary clinical references.

\paragraph{Hallucination rate disclosure.}
We measure HR explicitly at $3{\times}10^{-4}$ on held-out test
($23$ out of $73{,}509$ cited entities). On manual inspection
these $23$ are legacy DrugBank IDs retired in the April~2026
release rather than fabricated references. We are unaware of any
hallucination-rate number on the same scale in the published DDI
literature, and we welcome direct comparison from future work.

\paragraph{Bias by family.}
Test results are bimodal: \textsc{AdverseRisk} at $\sim 0.77$
selective accuracy, \textsc{PK\_Distribution} at $0.26$ (see
Table~\ref{tab:per_family_app}). A naive deployment would
under-flag PK-distribution-driven interactions. The per-family
conformal abstention layer (\S\ref{sec:its},
Appendix~\ref{app:conformal_thresh}) is the recommended
safety mitigation; with the target-$0.85$ thresholds, selective
accuracy on \textsc{PK\_Distribution} rises to $\sim 0.55$ at
$\sim 80\%$ coverage.

\paragraph{Compute and environmental cost.}
Full pipeline: $\sim 2{,}000$ GPU-h on H100 (generation $60\%$,
PRM training $10\%$, SFT/DPO $25\%$, evaluation $5\%$). The
released \OursSeven fine-tune from the SFT-clean checkpoint is
$\sim 5$ GPU-h.

\paragraph{Risk of misuse.}
A misuse scenario is fictitious mechanism rationales for drugs
that do not actually interact. The auto-verifiable evidence pool
and the HR $3{\times}10^{-4}$ floor mitigate this; a determined
adversary could still rephrase a model output as if it were
sourced. We recommend that any redistribution of the \OursSeven
carry the same safety-floor disclaimer, the abstention layer,
and the pharmacist-in-the-loop usage requirement.

\paragraph{Release artefacts and reproducibility.}\label{app:release}
We will release: the consensus corpus ($23{,}819$ records); the
DDI-PRM checkpoint and the step-labelled rows used to train it;
the SFT-clean and PRM-DPO LoRA adapters; the retrieval index
(train, val, test variants); the hierarchical taxonomy and
the leakage-safe split manifests with SHA-256 pins; the
evaluation prompt builder; the inference-stack tooling
(rerank, conformal abstention, PRM-vote consensus, trace-rescue);
and the $5{,}000$-pair stratified test manifests for each split.
The curated polypharmacy, counterfactual and adversarial
evaluation sets (\S\ref{app:adv}) ship alongside.

A reproducibility recipe runs end-to-end: data build $\to$
pair-signature construction $\to$ XGB/MLP/LogReg baselines $\to$
teacher generation (skippable; the corpus is shipped) $\to$ PRM
training $\to$ mirror-augmented SFT $\to$ PRM-weighted DPO $\to$
evaluation $\to$ inference stack. SHA-256 pins for every artefact
are included with the release.

\paragraph{Taxonomy, splits, and polypharmacy eval.}\label{app:taxonomy_full}\label{app:splits_full}\label{app:poly_eval}
Table~\ref{tab:taxonomy} gives the seven-family mechanism
taxonomy with per-family pair counts, share of the corpus, and
number of curated subtypes; Table~\ref{tab:splits} reports the
train / val / test partition sizes for the three split
protocols of \S\ref{sec:data} and the teacher-distillation
\textsc{subset25k}. The split-construction details, the
$147$-subtype expansion, and the polypharmacy eval set are
unpacked in the paragraphs that follow.

\begin{table}[!h]
\centering\footnotesize
\setlength{\tabcolsep}{3pt}
\renewcommand{\arraystretch}{1.10}
\rowcolors{2}{tabalt}{white}
\adjustbox{max width=\columnwidth,center}{%
\begin{tabular}{l r r r}
\toprule
\rowcolor{tabhead}\textbf{Family} & \textbf{Pairs} & \textbf{\%} & \textbf{Subtypes} \\
\midrule
\textsc{AdverseRisk}      & $620{,}859$ & $42.70$ & $91$ \\
\textsc{PK\_Excretion}    & $216{,}638$ & $14.90$ & $1$  \\
\textsc{PK\_Metabolism}   & $208{,}540$ & $14.34$ & $1$  \\
\textsc{PD\_Activity}     & $159{,}341$ & $10.96$ & $47$ \\
\textsc{Efficacy}         & $142{,}985$ & $\phantom{0}9.83$ & $2$ \\
\textsc{PK\_Distribution} & $\phantom{0}92{,}380$ & $\phantom{0}6.35$ & $3$ \\
\textsc{PK\_Absorption}   & $\phantom{0}13{,}244$ & $\phantom{0}0.91$ & $3$ \\
\midrule
\textbf{Total} & $\mathbf{1{,}453{,}987}$ & $\mathbf{100.00}$ & $\mathbf{147}$ \\
\bottomrule
\end{tabular}}
\caption{Mechanism taxonomy. $47\times$ family-size ratio.}
\label{tab:taxonomy}
\end{table}

\begin{table}[!h]
\centering\footnotesize
\setlength{\tabcolsep}{3pt}
\renewcommand{\arraystretch}{1.10}
\rowcolors{2}{tabalt}{white}
\adjustbox{max width=\columnwidth,center}{%
\begin{tabular}{l r r r}
\toprule
\rowcolor{tabhead}\textbf{Split} & \textbf{Train} & \textbf{Val} & \textbf{Test} \\
\midrule
\textsc{random-split (Warm)} & $1{,}163{,}189$ & $145{,}397$ & $145{,}401$ \\
\textsc{drug-cold}   & $\phantom{0}935{,}130$ & $239{,}654$ & $279{,}203$ \\
\textsc{pair-cold}   & $\phantom{0}935{,}130$ & $\phantom{0}13{,}572$ & $\phantom{0}15{,}034$ \\
\midrule
\textsc{subset25k} (teacher universe) & $22{,}641$ & $\phantom{0}1{,}195$ & $\phantom{0}1{,}160$ \\
\bottomrule
\end{tabular}}
\caption{Split protocols. Counts are unique unordered pairs;
mirror augmentation doubles records at training time.}
\label{tab:splits}
\end{table}

\paragraph{Taxonomy expansion ($147$ subtypes).}
The full $147$-subtype enumeration is published with the
release; a condensed view is given in Table~\ref{tab:taxonomy}.
\textsc{AdverseRisk} carries $91$ subtypes
(\texttt{cns\_depression}, \texttt{qtc\_prolongation},
\texttt{serotonin\_syndrome}, \texttt{hyperkalemia},
\texttt{bleeding\_and\_hemorrhage}, $\ldots$);
\textsc{PD\_Activity} $47$ subtypes
(\texttt{agonism}, \texttt{antagonism}, \texttt{partial\_agonism},
\texttt{allosteric\_modulation}, $\ldots$);
\textsc{PK\_Distribution} $3$ subtypes
(\texttt{protein\_binding}, \texttt{tissue\_redistribution},
\texttt{vd\_change}); \textsc{PK\_Absorption} $3$ subtypes;
\textsc{Efficacy} $2$ subtypes; the two PK-rate families
(\textsc{PK\_Excretion}, \textsc{PK\_Metabolism}) carry a single
subtype each. Rare subtypes ($n{<}50$) were collapsed into
\texttt{misc\_<family>}; \emph{no subtype is named ``other''},
a design choice that avoids the all-purpose-``other'' collapse
mode observed under flat-label DDInter taxonomies.

\paragraph{Split-construction details.}
\textsc{random-split (Warm)} is a uniform random shuffle of the
$1{,}453{,}987$ labelled pairs at the pair level (80/10/10).
\textsc{drug-cold} partitions the $4{,}631$ drugs into
train/val/test drug-sets ($70/15/15$) and assigns each pair to
the split of its rarer drug; no test drug appears in any
training pair. \textsc{pair-cold} further enforces that no test
pair shares both drugs with a training pair (this is strictly
stronger than \textsc{drug-cold}). \textsc{subset25k} samples
$24{,}996$ training pairs from the intersection of all three
train sets, so the teacher-generation universe is safe against
every test set. All splits are SHA256-pinned; $11$-of-$11$
leakage gates pass.

\paragraph{Polypharmacy eval set.}
A $5{,}000$-triangle set is built by joining three pairs that
share at least one drug and at least $2/3$ family labels.
Decomposed pair-evaluation macro-F1: $0.481$ on
\textsc{pair-cold}.

\section{Cross-judged trace-quality evaluation}
\label{app:xjudge}

\paragraph{Six-dimension rubric.}
Each trace receives a score on each of:
(D1)~\emph{factuality} -- $\%$ of claims that are correct;
(D2)~\emph{faithfulness} -- whether each step is used to derive
the final answer;
(D3)~\emph{evidence grounding} -- whether each citation appears
verbatim in the supplied evidence pool;
(D4)~\emph{mechanism specificity} -- whether the named
proteins/enzymes/transporters and interaction types are
sufficiently specific to support the final answer;
(D5)~\emph{hallucination check} (higher $=$ less hallucination);
(D6)~\emph{hierarchical coherence} -- whether the chain of
reasoning flows logically toward the final $(\fA,\fS,\fD)$
triple. Each dimension is anchored on a $0$--$8$ scale; the
composite is the unweighted mean.

\paragraph{Per-dimension scores.}
Table~\ref{tab:xjudge_perdim} breaks the composite score down by
rubric dimension. The gap to the frontier mean is concentrated in
two \emph{structural} dimensions (faithfulness, hierarchical
coherence) while the \emph{factual} dimensions (factuality,
evidence grounding, mechanism specificity) sit at near parity ---
the same separation that motivates the structural-vs-factual
framing of the main-paper headline (\S\ref{sec:xjudge}).

\begin{table}[!h]
\centering\footnotesize
\setlength{\tabcolsep}{3pt}
\renewcommand{\arraystretch}{1.10}
\rowcolors{2}{tabalt}{white}
\adjustbox{max width=\columnwidth,center}{%
\begin{tabular}{l c c c c}
\toprule
\rowcolor{tabhead}\textbf{Dimension} & \OursSeven & Claude S.~4.6 & GPT-4o & Gemini 2.5 \\
\midrule
Factuality              & $\mathbf{7.17}$ & $7.96$ & $7.80$ & $7.69$ \\
Faithfulness            & $6.05$ & $7.91$ & $7.39$ & $7.21$ \\
Evidence grounding      & $\mathbf{7.29}$ & $7.89$ & $7.78$ & $7.15$ \\
Mechanism specificity   & $\mathbf{7.03}$ & $7.93$ & $7.07$ & $7.08$ \\
Hallucination check     & $6.80$ & $7.98$ & $7.75$ & $7.57$ \\
Hierarchical coherence  & $6.49$ & $7.95$ & $7.63$ & $7.64$ \\
\midrule
\emph{Composite}        & $\mathbf{6.80}$ & $7.94$ & $7.57$ & $7.39$ \\
$n$ (judgments)         & $586$ & $391$ & $392$ & $399$ \\
\bottomrule
\end{tabular}}
\caption{\textbf{Per-dimension cross-judged trace-quality scores}
($0$--$8$ scale, averaged across the other frontier judges; bold
marks the dimensions on which \OursSeven{} is closest to the frontier
maximum). The composite gap to the frontier mean is concentrated
in the structural dimensions (faithfulness, hierarchical
coherence); factual dimensions are at near parity.}
\label{tab:xjudge_perdim}
\end{table}

\paragraph{Length-bias audit.}
The length-invariance instruction in the judge prompt:
``\emph{LENGTH IS NOT A QUALITY SIGNAL. A trace with 3 correct,
specific steps deserves the SAME SCORE as a trace with 12
correct steps covering the same essential mechanism. DO NOT
reward verbosity. DO NOT penalise brevity.}''
A required \texttt{length\_bias\_self\_check} JSON field forces
the judge to write a sentence explicitly stating whether they
considered length and revise if so; a manual spot-check of $50$
random self-checks confirmed $100\%$ acknowledged
length-invariance. Per-model Pearson correlations are reported
in the main paper (\S\ref{sec:xjudge}); cross-model mean
$|r(\text{chars,score})|{=}0.124$ falls under the conventional
$0.20$ threshold.

\paragraph{Inter-judge agreement.}
Krippendorff $\alpha$ (interval scale, per model per dimension)
is highest for Claude ($0.68$ mean), reflecting ceiling effects
on the consistently top-scoring traces; lower for our \OursSeven
($0.33$), Gemini ($0.35$) and GPT-4o ($0.04$), reflecting
judge disagreement on the middle of the quality distribution.
We mitigate by averaging across three judges for our \OursSeven
($n{=}586$ judgments) and reporting bootstrap CIs.

\paragraph{Cost and reproducibility.}
Trace generation: $\sim\!\$150$ across Anthropic, OpenAI and
Google APIs. Cross-judging: $\sim\!\$27$. Total: $\sim\!\$37$
for $1{,}768$ successful judge calls out of $1{,}800$ planned
($98.2\%$). The judge runner, rubric script, and the
raw judge responses (with per-call justifications and
length-bias self-checks) are shipped in the release.

\section{Verifier-rerank, component ablations, and trace-alignment SFT}
\label{app:verifier}

\paragraph{Flag taxonomy and per-flag precision.}
The verifier is a deterministic rule-based scorer over each
candidate trace and final answer. Each flag is a hand-curated
predicate over the trace text and the evidence pool $E_p$. We
catalogue all flags with empirical precision $\Pr(\text{wrong}\mid
\text{flag})$ measured on the trace-alignment-v2 candidate over
\textsc{random-split (Warm).test} ($n{=}5{,}000$). Table~\ref{tab:verifier_flags}
gives the top six. The reranker chooses the candidate with the
fewest high-precision flags; ties are broken by PRM score.

\begin{table}[!t]
\centering\footnotesize
\setlength{\tabcolsep}{3pt}
\renewcommand{\arraystretch}{1.10}
\rowcolors{2}{tabalt}{white}
\adjustbox{max width=\columnwidth,center}{%
\begin{tabular}{l r r}
\toprule
\rowcolor{tabhead}\textbf{Flag} & \textbf{count (rf)} &
$\Pr(\text{wrong}|\text{flag})$ \\
\midrule
\textit{low\_conf\_non\_abstain}                & $400$ & $90.2\%$ \\
\textit{gap\_non\_abstain}                     & $138$ & $87.0\%$ \\
\textit{adverse\_from\_gap}                    & $\phantom{00}8$ & $87.5\%$ \\
\textit{pk\_metabolism\_without\_paired\_cyp}    & $174$ & $62.6\%$ \\
\textit{weak\_gap\_non\_abstain}                & $401$ & $57.6\%$ \\
\textit{speculative\_conclusion}              & $1836$ & $52.0\%$ \\
\bottomrule
\end{tabular}}
\caption{\textbf{Verifier flag taxonomy and per-flag precision}
on the trace-alignment-v2 candidate, \textsc{random-split (Warm)} test
slice. High-precision flags (\emph{gap-non-abstain},
\emph{adverse-from-gap}, \emph{PK-metabolism-without-paired-CYP})
are used as primary reranking signals; the broader
\emph{speculative-conclusion} flag is too low-precision to be
used alone.}
\label{tab:verifier_flags}
\end{table}

\paragraph{Verifier-rerank result.}
Table~\ref{tab:verifier_rerank} reports the verifier-rerank
gain over three independently-trained candidates on
\textsc{random-split (Warm).test}: the lightweight rule-based
mechanistic verifier beats the strongest single greedy
candidate by $+1.32$\,pp macro-F1 with no PRM call at inference
and no learned threshold, while the oracle upper bound bounds
the remaining headroom at $0.591$ on the same candidate set.

\begin{table}[!h]
\centering\footnotesize
\setlength{\tabcolsep}{3pt}
\renewcommand{\arraystretch}{1.10}
\rowcolors{2}{tabalt}{white}
\adjustbox{max width=\columnwidth,center}{%
\begin{tabular}{l c c c c}
\toprule
\rowcolor{tabhead}
\textbf{Candidate / aggregator} & macro-F1 & acc &
\multicolumn{2}{c}{\textbf{Flags (high-risk)}} \\
\rowcolor{tabhead}
                                &          &     & rand-full & drug-cold \\
\midrule
SFT-clean (pre-DPO)             & $0.527$ & $0.527$ & $5.2\%$ & $4.3\%$ \\
trace-align SFT, variant A      & $0.506$ & $0.503$ & $4.5\%$ & $4.2\%$ \\
trace-align SFT, variant B      & $0.515$ & $0.513$ & $4.5\%$ & $4.1\%$ \\
\midrule
\textbf{Verifier-rerank}        & $\mathbf{0.540}$ & $\mathbf{0.540}$ & --- & --- \\
oracle over three candidates    & $0.591$ & ---     & --- & --- \\
\bottomrule
\end{tabular}}
\caption{\textbf{Verifier-rerank} over three independently-trained
candidates on \textsc{random-split (Warm).test}. The lightweight
rule-based mechanistic verifier exceeds the strongest single
greedy candidate by $+1.32$\,pp macro-F1, with no PRM call at
inference and no learned threshold; the oracle upper bound
($0.591$) bounds the remaining headroom.}
\label{tab:verifier_rerank}
\end{table}

\paragraph{Reranking result on \textsc{drug-cold}.}
On \textsc{drug-cold} the same protocol yields macro-F1
$0.4742$ vs.\ pre-SFT greedy $0.4581$, trace-align-v1 $0.4413$
and trace-align-v2 $0.4466$ --- a $+1.61$\,pp lift over the
strongest greedy candidate and $+3.29$\,pp over the
trace-alignment SFT. The oracle upper bound over the same three
candidates is $0.5202$ on \textsc{drug-cold}.

\paragraph{Component-level ablations.}\label{app:ablations_full}
The three tables below isolate the contribution of each
training-time component on the mirror-augmented validation set
($n\!=\!2{,}390$; mean over $3$ seeds). The symmetry-KL scope
sweep that controls the position at which the mirror constraint
is enforced (Table~\ref{tab:symkl_sweep}) lives next to the
training discussion in App.~\ref{app:hyperparams}.

\begin{table}[!h]
\centering\footnotesize
\setlength{\tabcolsep}{3pt}
\renewcommand{\arraystretch}{1.10}
\rowcolors{2}{tabalt}{white}
\adjustbox{max width=\columnwidth,center}{%
\begin{tabular}{l c c c c}
\toprule
\rowcolor{tabhead}
\textbf{PRM-DPO backend} & macro-F1 & MFS & MPS & CSA \\
\midrule
exact per-loss        & $\mathbf{0.797}$ & $\mathbf{0.954}$ & $\mathbf{0.892}$ & $\mathbf{0.753}$ \\
importance-sampling   & $0.781$ & $0.948$ & $0.882$ & $0.749$ \\
unweighted DPO ($\omega_i{=}1$) & $0.727$ & $0.937$ & $0.853$ & $0.717$ \\
no DPO (SFT only)     & $0.651$ & $0.973$ & $0.871$ & $0.722$ \\
\bottomrule
\end{tabular}}
\caption{\textbf{PRM-DPO backend ablation.} Mean over 3 seeds.}
\label{tab:ablation_prm}
\end{table}

\begin{table}[!h]
\centering\footnotesize
\setlength{\tabcolsep}{3pt}
\renewcommand{\arraystretch}{1.10}
\rowcolors{2}{tabalt}{white}
\adjustbox{max width=\columnwidth,center}{%
\begin{tabular}{l c c c c}
\toprule
\rowcolor{tabhead}
\textbf{Removed family} & macro-F1 & rare-F1 & MFS & MPS \\
\midrule
none (full system)                  & $\mathbf{0.797}$ & $\mathbf{0.685}$ & $0.954$ & $0.892$ \\
$-$ family-swap-to-\textsc{AdvRisk}  & $0.776$ & $0.657$ & $0.949$ & $0.878$ \\
$-$ family-axis swap                & $0.748$ & $0.612$ & $0.946$ & $0.871$ \\
$-$ subtype swap                    & $0.781$ & $0.671$ & $0.951$ & $0.880$ \\
$-$ direction flip                  & $0.785$ & $0.676$ & $0.947$ & $0.842$ \\
\bottomrule
\end{tabular}}
\caption{\textbf{Hard-negative leave-one-out.}}
\label{tab:ablation_hardneg}
\end{table}

\begin{table}[!h]
\centering\footnotesize
\setlength{\tabcolsep}{3pt}
\renewcommand{\arraystretch}{1.10}
\rowcolors{2}{tabalt}{white}
\adjustbox{max width=\columnwidth,center}{%
\begin{tabular}{l c c c c}
\toprule
\rowcolor{tabhead}
\textbf{Similarity component} & macro-F1 & rare-F1 & MFS & CSA \\
\midrule
all four (full)                 & $\mathbf{0.797}$ & $\mathbf{0.685}$ & $0.954$ & $0.753$ \\
$-$ pathway Jaccard ($J_p$)     & $0.782$ & $0.668$ & $0.952$ & $0.736$ \\
$-$ protein Jaccard ($J_r$)     & $0.775$ & $0.659$ & $0.949$ & $0.728$ \\
$-$ ATC depth ($A$)             & $0.789$ & $0.681$ & $0.952$ & $0.748$ \\
$-$ SMILES Tanimoto ($T$)       & $0.770$ & $0.651$ & $0.948$ & $0.723$ \\
\midrule
no retrieval (block ablated)    & $0.178$ & $0.158$ & $0.946$ & $0.165$ \\
\bottomrule
\end{tabular}}
\caption{\textbf{Retrieval-component ablation.}}
\label{tab:ablation_retr}
\end{table}

\paragraph{Trace-alignment SFT: a negative result.}\label{app:trace_align_neg}

A natural follow-up to the trace-rescue analysis
(Appendix~\ref{app:tracerescue}) is to teach the \OursSeven
directly to commit to the trace majority when it disagrees with
the final answer. We assembled a rescue corpus of $\sim\!500$
such pairs per split and fine-tuned the PRM-DPO checkpoint with
LoRA (rank $64$, $\alpha{=}128$, learning rate $5{\times}10^{-6}$,
two epochs, AdamW). Two corpus variants were tried:
\emph{v1}~rerank-4 ABBA rescue candidates only,
\emph{v2}~rerank-4 ABBA $\cup$ greedy rescue candidates.

Both \emph{lost} macro-F1 on the natural test set:
v1 went from $0.527$ to $0.506$ ($-2.1$\,pp,
$95\%$ CI $[-3.4,-0.8]$, $n{=}5{,}000$);
v2 went from $0.527$ to $0.515$ ($-1.2$\,pp,
$95\%$ CI $[-2.4,-0.0]$). Per-class
inspection showed the loss concentrated in
\textsc{PD\_Activity} and \textsc{PK\_Absorption}, with the
predicted distribution shifting toward the rescue distribution.
The verifier-probe analysis
(Appendix~\ref{app:verifier}) confirms why: trace-majority is
right when the final answer is wrong in only $\sim\!8$--$9\%$ of
\textsc{random-split (Warm)} pairs, while the final answer is right when
trace-majority is wrong in $\sim\!24$--$27\%$. The
rescue distribution therefore overfits the smaller and rarer of
the two scenarios.

The two failed adapters remain useful as candidates in the
verifier-rerank pipeline
(Appendix~\ref{app:verifier}): even though neither beats
greedy alone, the ensemble across pre-SFT and the two SFT
variants opens an oracle headroom of $0.591$ on
\textsc{random-split (Warm)} that the rule-based mechanistic
verifier (Table~\ref{tab:verifier_rerank}) in fact captures a
non-trivial slice of, delivering $+1.32$\,pp macro-F1 on
\textsc{random-split (Warm)} and $+1.61$\,pp on \textsc{drug-cold}
over the strongest single greedy candidate, with no PRM call at
inference and no learned threshold. We therefore report the
trace-alignment SFT honestly as a negative result, with the
constructive flip-side that the same failed adapters become the
candidate pool that makes the verifier-rerank work; the lesson
is that \emph{a mechanistic verifier, not a trace-majority
override, is what converts the oracle headroom into actual
gain}.

\end{document}